\documentclass[preprint,12pt,authoryear]{elsarticle}

\usepackage{amssymb}
\usepackage{hyperref}
\usepackage{booktabs}
\usepackage{rotating}

\usepackage{amsmath,amsfonts}
\usepackage{algorithmic}
\usepackage{array}
\usepackage{subfig}
\usepackage{textcomp}
\usepackage{stfloats}
\usepackage{url}
\usepackage{verbatim}
\usepackage{graphicx}
\usepackage{float}

\usepackage{mathtools}
\usepackage{tabularx}
\usepackage{multirow}
\usepackage{adjustbox}
\usepackage{epsfig}
\usepackage{boxhandler}
\usepackage{hhline}
\usepackage{balance}

\usepackage{xcolor}%

\journal{Engineering Applications of Artificial Intelligence}

\begin{document}

\begin{frontmatter}


\title{ESRPCB: an Edge guided Super - Resolution model and Ensemble learning for tiny Printed Circuit Board Defect detection} 

\author[1]{Xiem HoangVan} \corref{cor1} 
\ead{xiemhoang@vnu.edu.vn}

\author[1]{Dang Bui Dinh}
\ead{dangdinh1713@gmail.com}

\author[2]{Thanh Nguyen Canh}
\ead{thanhnc@jaist.ac.jp}

\author[3]{Van-Truong Nguyen}
\ead{qvtruongcdt@gmail.com}

\cortext[cor1]{Corresponding author}

\affiliation[1]{organization={Faculty of Electronics and Telecommunications, University of Engineering and Technology, Vietnam National University},
           addressline={Cau Giay}, 
           city={Hanoi},
           postcode={10000}, 
           country={Vietnam}}
           
\affiliation[2]{organization={School of Information Science, Japan Advanced Institute of Science and Technology},
           addressline={Nomi}, 
           city={Ishikawa},
           postcode={923-1211}, 
           country={Japan}}
\affiliation[3]{organization={Faculty of Mechatronics, SMAE, Hanoi University of Industry},
           addressline={Cau Dien}, 
           city={Hanoi},
           postcode={10000}, 
           country={Vietnam}}

\begin{abstract}
Printed Circuit Boards (PCBs) are critical components in modern electronics, which require stringent quality control to ensure proper functionality. However, the detection of defects in small-scale PCBs images poses significant challenges as a result of the low resolution of the captured images, leading to potential confusion between defects and noise. To overcome these challenges, this paper proposes a novel framework, named ESRPCB (edge-guided super-resolution for PCBs defect detection), which combines edge-guided super-resolution with ensemble learning to enhance PCBs defect detection. The framework leverages the edge information to guide the EDSR (Enhanced Deep Super-Resolution) model with a novel ResCat (Residual Concatenation)  structure, enabling it to reconstruct high-resolution images from small PCBs inputs. By incorporating edge features, the super-resolution process preserves critical structural details, ensuring that tiny defects remain distinguishable in the enhanced image. Following this, a multi-modal defect detection model employs ensemble learning to analyze the super-resolved image, improving the accuracy of defect identification. Experimental results demonstrate that ESRPCB achieves superior performance compared to State-of-the-Art (SOTA) methods, achieving an average Peak Signal to Noise Ratio (PSNR) of $30.54 $ $dB(decibel)$, surpassing EDSR by $0.42 dB$. In defect detection, ESRPCB achieves a mAP50(mean average precision at an Intersection over Union threshold of 0.50) of $0.965$, surpassing EDSR ($0.905$) and traditional super-resolution models by over $5\%$. Furthermore, the ensemble-based detection approach further enhances performance, achieving a mAP50 of $0.977$. These results highlight the effectiveness of ESRPCB in enhancing both image quality and defect detection accuracy, particularly in challenging low-resolution scenarios.

\end{abstract}


\end{frontmatter}

\section{Introduction} \label{sec:Intro}

Printed Circuit Boards (PCBs) form the foundation of modern electronic systems, serving as essential platforms for electrical interconnections between components. As the core building block of any electronic design, PCBs have evolved into highly sophisticated components with superior quality over the years, with ongoing advancements in miniaturization and performance. However, as PCBs become increasingly compact, the challenge of detecting defects during manufacturing becomes more critical. Defects in PCBs can lead to severe malfunctions, increase production costs, and compromise product reliability, making early and accurate defect detection crucial in maintaining quality standards during manufacturing~(\cite{putera2010printed}).

\begin{figure}[htbp] 
        \captionsetup[subfloat]{labelformat=empty}
        \begin{center}
            \newcommand{\rowArg}{2.65cm}
            \newcommand{\patchsize}{3.4cm}
            \setlength\tabcolsep{6pt}
            \begin{tabular}{c c c}
                \subfloat[Missing hole]{\includegraphics[width=\patchsize]{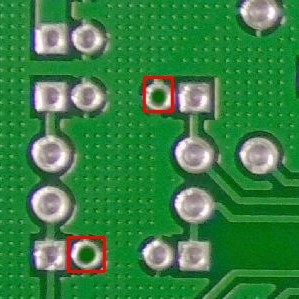}} &
                \subfloat[Mouse bite]{\includegraphics[width=\patchsize]{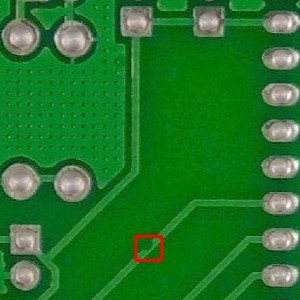}} &
                \subfloat[Open circuit]{\includegraphics[width=\patchsize]{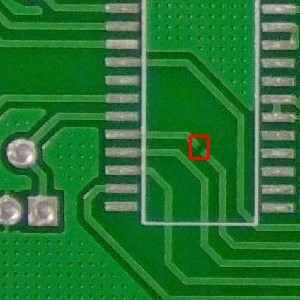}} \\ [0cm]
                
                \subfloat[Short]{\includegraphics[width=\patchsize]{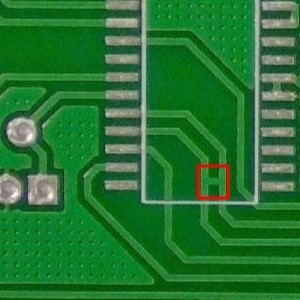}} &
                \subfloat[Spur]{\includegraphics[width=\patchsize]{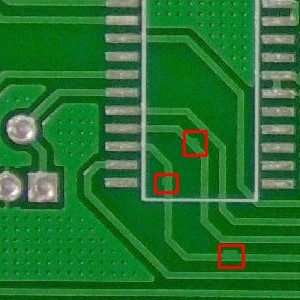}} &
                \subfloat[Spurious copper]{\includegraphics[width=\patchsize]{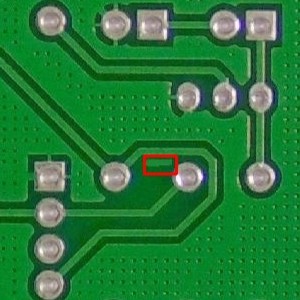}} \\
            \end{tabular}
        \end{center}
        \vspace*{-0.3cm}
        \caption{Example of PCB images that contain defects.}
        \label{fig:six_defect}
    \end{figure}

PCB defects can generally be categorized into two main types: functional defects and cosmetic defects~(\cite{tddnet}). Functional defects directly impact the performance of the PCB, potentially leading to malfunctions or complete failures in electronic devices. As such, these are the most critical defects to detect and rectify during the manufacturing process. On the other hand, cosmetic defects, while seemingly limited to visual imperfections, can cause long-term performance degradation due to issues such as abnormal heat dissipation or uneven current distribution. Both types of defects are important for maintaining the reliability and lifespan of PCBs in modern electronics. Among these, there are six commonly encountered defects in industrial settings: missing hole, mouse bite, open circuit, short, spur, and spurious copper. These defects are illustrated in Fig.~\ref{fig:six_defect} considered in this paper.

Several factors significantly affect the performance of PCB defect detection systems. First, the intricate and diverse structures and designs of PCBs, along with their complex design rules, pose challenges in developing universal algorithms that are compatible across different PCB layouts. This leads to instability in traditional defect detection methods. In addition, the types and characteristics of PCB defects vary significantly, and the low defect rate in industrial production complicates the collection of large numbers of defect samples, resulting in data imbalance that affects the performance of traditional methods~(\cite{tddnet}).

Environmental factors such as camera quality, lighting conditions, and PCB positioning further complicate defect detection. In industrial settings, where automated quality control systems rely on high-speed imaging, low-resolution images with noise, motion blur, and inconsistent lighting can hinder accurate defect identification. These factors make it essential to develop automated defect detection models that are robust to variations in image quality and capable of distinguishing between actual defects and noise~(\cite{AReviewandAnalysisofAOI}).

Traditional methods of PCB defect detection, such as manual inspection or rule-based image processing techniques, struggle to keep pace with the increasing complexity and miniaturization of PCBs in modern electronics. Manual inspections are not only time-consuming and labor-intensive but also prone to human error, especially in high-volume manufacturing environments. Rule-based methods, while automated, often fail to scale effectively for small-scale PCB images due to the low resolution and noise present in the images. Distinguishing between actual defects and noise becomes significantly more difficult as the quality of the images decreases, leading to a reduction in detection accuracy. These limitations are especially pronounced in high-density PCBs, where fine details are crucial for identifying potential defects. As a result, there is a growing need for advanced, automated detection techniques capable of overcoming the challenges posed by low-resolution imaging and high defect variability.

In recent years, the use of machine vision techniques has revolutionized PCB defect detection. Vision-based methods such as the template matching method and the OpenCV-based image subtraction method have been used to detect PCB defects~(\cite{reviewimageprocessing}). Unfortunately, these methods are vulnerable to variations in transition and rotation, and they are limited to specific defect types, often failing to detect missing or undersized components. The Canny edge detection algorithm is a widely used edge detection technique for identifying defects on PCBs and performs well in low-light conditions. However, it has drawbacks, including complex computations and sensitivity to noise. The reference-free path-walking method is another useful technique for detecting PCB faults without a reference design. However, its accuracy diminishes if measurements are not performed precisely, and it is time-consuming. X-ray computed tomography (XCT) is a widely used method to detect PCB defects, since it uses X-rays to create cross-sectional images, allowing for a more detailed examination of the PCB structure~(\cite{AutomaticPCBinspectionsystems}). Nevertheless, this method has several limitations, including high costs, time consumption, and complexity in evaluating image results~(\cite{PCBdd_survey}),~(\cite{surveydefectdetection}). 

Early approaches focused on leveraging convolutional neural networks (CNNs) for defect detection in PCBs, which yielded promising results but struggled with low-resolution images, especially when detecting defects in tiny or intricate components of PCBs. State-of-the-art approaches have sought to overcome the limitations of traditional image resolution and noise. For example, Zhang \textit{et al.}~(\cite{zhang2018improved}) introduced a You Only Look Once (YOLO)v3-based defect detection framework, which demonstrated high-speed detection but struggled with very small defects in low-resolution images. Chen \textit{et al.}~(\cite{chen2023pcb}) extended this by incorporating a deep learning model that focused on enhancing image quality for PCB defect detection. Furthermore, study~(\cite{tang2019online}) presented a dataset-driven method that applied deep learning for defect classification but highlighted the need for improved resolution in detecting small defects. Recent advancements such as the YOLOv8 model and Enhanced Deep Super-Resolution (EDSR)~(\cite{lim2017enhanced}) networks have shown promise in addressing these issues by improving detection performance through better image reconstruction and higher detection accuracy.

Despite these advancements, the detection of tiny defects in low-resolution PCB images remains a significant challenge. Low-quality images, often distorted by noise, poor lighting, or compression artifacts, can lead to false positives or undetected defects. While many existing approaches utilize CNN-based models for classification, these methods frequently neglect the edge, which is critical for distinguishing noise from genuine defects. To overcome these challenges, a novel framework is proposed that integrates edge-guided super-resolution techniques with the EDSR networks to reconstruct high-resolution from low-quality inputs. Additionally, an ensemble learning-based detection model is incorporated to combine predictions from multiple classifiers, improving robustness and accuracy across diverse defect types and imaging conditions. 

The key contributions of this paper are as follows:

\begin{itemize}
    \item An edge-guided super-resolution framework: This novel approach incorporates edge information into the EDSR model with a novel Residual Concatenation (ResCat) structure, enabling the reconstruction of high-resolution images from small-scale, low-quality PCB inputs, thereby improving the visibility of tiny defects.
    \item Ensemble learning for defect detection: By combining predictions from multiple detection models, the ensemble learning framework enhances the accuracy and robustness of defect detection, especially for complex and varied defect scenarios in enhanced PCB images.
    \item Comprehensive evaluation on a standard PCB dataset: Extensive experiments demonstrate significant improvements in defect detection accuracy across various imaging conditions, validating the robustness and scalability of the proposed framework.
\end{itemize}

The remainder of this paper is structured as follows: Section~\ref{sec:relate work} reviews the related work in PCB defect detection and super-resolution techniques, providing insights into the challenges and advancements in these areas. Section~\ref{sec:proposed method} introduces the proposed framework, detailing the integration of edge-guided super-resolution with EDSR and the ensemble learning-based defect detection model. Section~\ref{sec:Eval} describes the experimental setup, dataset, and evaluation metrics, followed by a discussion of the results and comparative analysis. Finally, Section~\ref{sec:conclusion} concludes with a summary of the findings, the impact of the proposed approach, and directions for future research.

\section{Related Work} ~\label{sec:relate work}
\subsection{PCBs Defect Detection}
Detecting defects in Printed Circuit Boards is an essential task in industrial quality control, aimed at identifying various types of flaws on PCBs. Over the years, various methods have been developed to identify defects, ranging from traditional manual inspection and machine learning techniques to advanced deep learning-based approaches.

Early PCB defect detection methods relied heavily on manual inspection or mechanical methods, while widely adopted, suffered from slow speed and high costs, making them unsuitable for modern high-speed production environments~(\cite{wu2024eemnet}). These methods often struggled with scalability, especially as PCB designs became more complex and miniaturized. To improve efficiency, traditional image processing techniques, such as phase correlation~(\cite{hagi2014defect}), local binary patterns (LBP)~(\cite{lu2018defect}), and histogram of oriented gradients (HOG)~(\cite{lin2020surface}), were applied to PCB images. Although these methods provided reasonable accuracy in controlled scenarios, their reliance on manually engineered features limited their applicability in more complex and dynamic environments, where defects may vary significantly in appearance.

The advent of deep learning has led to significant advancements in PCB defect detection. Convolutional neural networks (CNNs) have become the dominant approach, as they can automatically learn relevant features from raw image data without the need for manual feature engineering. CNN-based methods are generally categorized into two types: one-stage and two-stage networks. One-stage detection models, such as You Only Look Once ~(\cite{yolo, redmon2017yolo9000}) and Single Shot MultiBox Detector (SSD)~(\cite{liu2016ssd}), offer faster processing speeds by performing object classification and localization in a single pass through the network. These models have gained popularity due to their ability to achieve real-time performance while maintaining relatively high accuracy. To address the issues of poor stability and low accuracy in PCB defect detection, various research~(\cite{jiang2022multi, li2022research, zheng2022printed, lim2023deep, zhao2022improved,xuan2022lightweight}) utilized a single network with many improvements including incorporation with the attention mechanism, path-aggregation feature, or replace backbone. The lightweight architecture of these models makes them particularly well suited for deployment on resource-constrained devices, such as embedded systems or industrial robots, where both speed and efficiency are critical.

In contrast, two-stage detection models, such as Region-based Convolutional Neural Networks (R-CNN)~(\cite{girshick2014rich}) and its derivatives like Fast R-CNN~(\cite{girshick2015fast}), Faster R-CNN~(\cite{ren2016faster}), and Mask R-CNN~(\cite{he2017mask}), first generate region proposals and then refine these regions to detect defects. These models have demonstrated high detection accuracy, especially in challenging scenarios where precise localization is critical. However, the complexity of two-stage models, particularly the inclusion of the region proposal network (RPN), leads to longer processing times and higher computational demands, making them unsuitable for real-time detection in industrial settings. Moreover, the large memory requirements and processing power needed for these models make it difficult to deploy them on low-power embedded systems, such as those used in mobile devices or edge computing.

One of the major challenges in PCB defect detection is accurately identifying tiny defects, which often have small pixel sizes, limited contextual information, and occur in complex backgrounds. Tiny defects can be particularly difficult to detect due to their subtle appearance and the noise introduced by the surrounding environment. To address this, several approaches have been developed such as study~(\cite{chen2017automatic}) proposed the atrous spatial pyramid pooling-balanced feature pyramid network (FPN) and studies~(\cite{zhang2023idd, jiang2023pcb}) focused on improving feature aggregation and integrating attention mechanisms to increase the model's ability to detect small, hard-to-spot defects. Zhou \textit{et al.}~(\cite{zhou2024efficient}) develops a compact defect detection model called Tiny Defect Detection YOLO (TDD-YOLO) and introduces a novel compression training methodology enabling training on low-resolution images while testing on those in their original resolution. However, TDD-YOLO sacrifices some detection precision in highly complex PCB layouts. Additionally, methods like generative adversarial networks (GANs) have been explored for tiny defect detection, with GAN-generated super-resolution images improving the quality of input images and thus enhancing detection accuracy.

Despite the improvements made by deep learning-based models, detecting tiny defects remains a challenging problem, especially when dealing with complex backgrounds that obscure the defects. Some researchers have employed coarse-to-fine detection approaches, where a coarse classifier is used to filter out irrelevant background regions before a fine classifier is applied to localize and classify the defects more accurately. Yang \textit{et al.}~(\cite{yang2019prior}) applied this method to detect tiny surface defects on aircraft engine blades, achieving better detection accuracy at the cost of increased computational complexity. While effective, these two-stage approaches may not meet the real-time processing requirements of modern industrial applications.

In addition to CNN-based models, Transformer-based architectures have recently been explored for PCB defect detection. Transformers, originally developed for natural language processing, have shown great promise in computer vision tasks due to their ability to capture long-range dependencies and model global context. Transformer-based models, such as the Vision Transformer (ViT) and Swin-Transformer, have been applied to PCB defect detection with promising results. An \textit{et al.}~(\cite{an2022lpvit}) introduced a label-robust and patch correlation-enhanced ViT (LPViT), which leveraged relationships between different regions of the PCB image to improve detection accuracy. However, transformer models typically require more computational resources and larger datasets compared to CNNs, making them less suitable for real-time industrial applications, particularly in resource-constrained environments. 

Recently, Generative Adversarial Networks (GANs) and Diffusion models have been explored to enhance PCB defect detection~(\cite{gulsuna2024design, patel2024musap, xu2025printed, xie2024mfad}) by generating synthetic high-quality defect samples to mitigate data imbalance and improve training efficiency. For instance,~(\cite{huang2025defect}) introduced a GAN-based defect augmentation framework to synthesize realistic PCB defect images, thereby addressing the challenge of limited defect samples in industrial datasets. Their method improved model generalization but suffered from mode collapse issues, where the generated samples lacked sufficient diversity, limiting robustness in real world applications. While their approach demonstrated superior defect localization, it required extensive computational resources due to the iterative nature of diffusion models, making it unsuitable for real-time manufacturing lines.~\cite{fang2025human} developed a GAN-diffusion fusion model, where GANs provided initial defect augmentation and diffusion models refined fine-grained textures. Despite achieving state-of-the-art accuracy, their approach exhibited slow convergence, necessitating careful hyperparameter tuning and extended training times. Additionally, the reliance on high-quality reference images limited its effectiveness when applied to real-world PCB images with severe noise and occlusions.

In summary, PCB defect detection has seen substantial advancements with the introduction of deep learning, particularly CNN-based methods. While two-stage models like Faster R-CNN provide high accuracy, they are often too slow and computationally intensive for real-time industrial applications. One-stage models, such as YOLO and SSD, offer a more practical solution, balancing speed and accuracy, making them ideal for real-world deployments. However, detecting tiny defects, especially in complex backgrounds, remains a challenge that continues to be a focus of ongoing research. Newer methods, such as Transformer-based models, have shown potential to further enhance detection accuracy, but their high computational requirements may limit their use in real-time settings. The future of PCB defect detection likely lies in the development of hybrid approaches that combine the strengths of CNNs, Transformers, and attention mechanisms to improve both the accuracy and efficiency of defect detection systems.

\subsection{Super-Resolution}

Super-resolution (SR) techniques have become an essential tool in image processing, aiming to enhance the resolution of low-quality images by reconstructing finer details. This is particularly valuable in industrial applications like printed circuit board (PCB) defect detection, where image clarity and detail are critical for accurately identifying defects, especially when the defects are tiny or hidden within complex textures. With advancements in deep learning, super-resolution models have evolved to significantly outperform traditional approaches, offering better results in terms of both detail recovery and processing speed.

Early approaches to super-resolution relied heavily on interpolation methods, such as bicubic~(\cite{keys1981cubic}) or bilinear~(\cite{kirkland2010bilinear}) interpolation, which increase image resolution by estimating pixel values based on surrounding data points. However, these methods often resulted in blurry outputs and failed to recover fine details, particularly in highly textured or complex images. More sophisticated techniques, such as example-based super-resolution~(\cite{freeman2002example}), improved upon interpolation methods by leveraging large image databases to reconstruct high-frequency details in low-resolution images. However, these methods required significant computational resources and were constrained by the necessity for large training datasets.

The advent of deep learning significantly improved SR, with models like Enhance Deep Super-Resolution (EDSR)~(\cite{lim2017enhanced}) becoming widely adopted for their ability to reconstruct high-frequency details more effectively than earlier methods. By removing batch normalization layers, EDSR focuses on minimizing artifacts and improving image clarity, making it especially useful in industrial applications such as PCB defect detection. Enhancing image resolution helps in identifying small and subtle defects like micro-cracks or misaligned components, which are often missed in lower-resolution images.  Additionally, EDSR's lightweight architecture ensures it operates efficiently in real-time systems, which is essential for automated manufacturing environments.

While other models like Generative Adversarial Networks (GANs) and their variants, such as ESRGAN~(\cite{wang2018esrgan}), BSRGAN~(\cite{zhang2021designing}) and Real-ESRGAN~(\cite{wang2021real}), have also shown promise in generating high-quality super-resolved images, they tend to require more computational resources and can be prone to instability during training. In contrast, EDSR strikes a balance between computational efficiency and image quality, making it a preferred choice for enhancing PCB images.

The integration of SR models like EDSR into PCB defect detection workflows has been explored, where enhanced images are passed into detection algorithms to improve accuracy. This combination allows for more reliable identification of tiny defects, ensuring greater fidelity in the quality control process. As SR techniques continue to evolve, their application in industrial scenarios like PCB inspection is expected to further improve detection performance and efficiency.
\section{Proposed Method} ~\label{sec:proposed method}
\begin{figure}[htbp]
    \centering
    \includegraphics[width=0.8\linewidth]{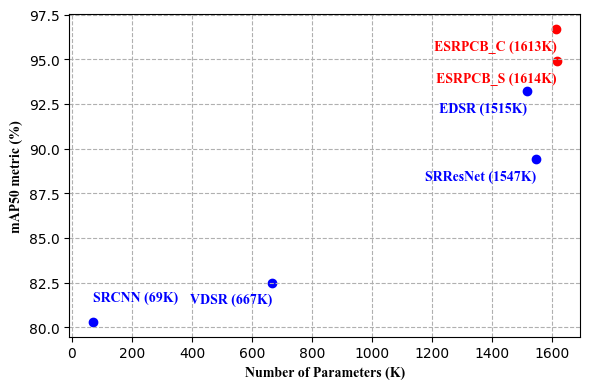}
    \caption{Trade-off between accuracy metrics (mAP50) and model parameters on the PCB Defect dataset with various SR models and the multimodal detection model. Our method is highlighted in red.}
    \label{fig:tradeoff}
\end{figure}

This section presents a novel super-resolution model for a PCB defect detection system, named Edge guided Super-Resolution (ESRPCB) network. The proposed ESRPCB framework consists of two main components: a deep super-resolution model enhanced by edge information integration and a novel ResCat structure. These innovations aim to optimize the super-resolution process and improve defect detection accuracy. Additionally, the enhanced images will be processed by a multi-modal defect detection model based on the latest YOLO frameworks, maximizing the system's ability to detect faults accurately. Extensive experimental evaluations on the PCB Defect dataset demonstrate the superior performance of the ESRPCB network in terms of image quality reconstruction and defect detection, as illustrated in Fig.~\ref{fig:tradeoff}. To further investigate the impact of different edge detection techniques, two versions of our model are implemented: 1) ESRPCB\_C, which integrates edge information using the Canny algorithm, and 2) ESRPCB\_S, which utilizes the Sobel algorithm for edge enhancement.


\subsection{Our proposed architecture}

\begin{figure*}[!ht]
    \centering
    \includegraphics[width=\textwidth]{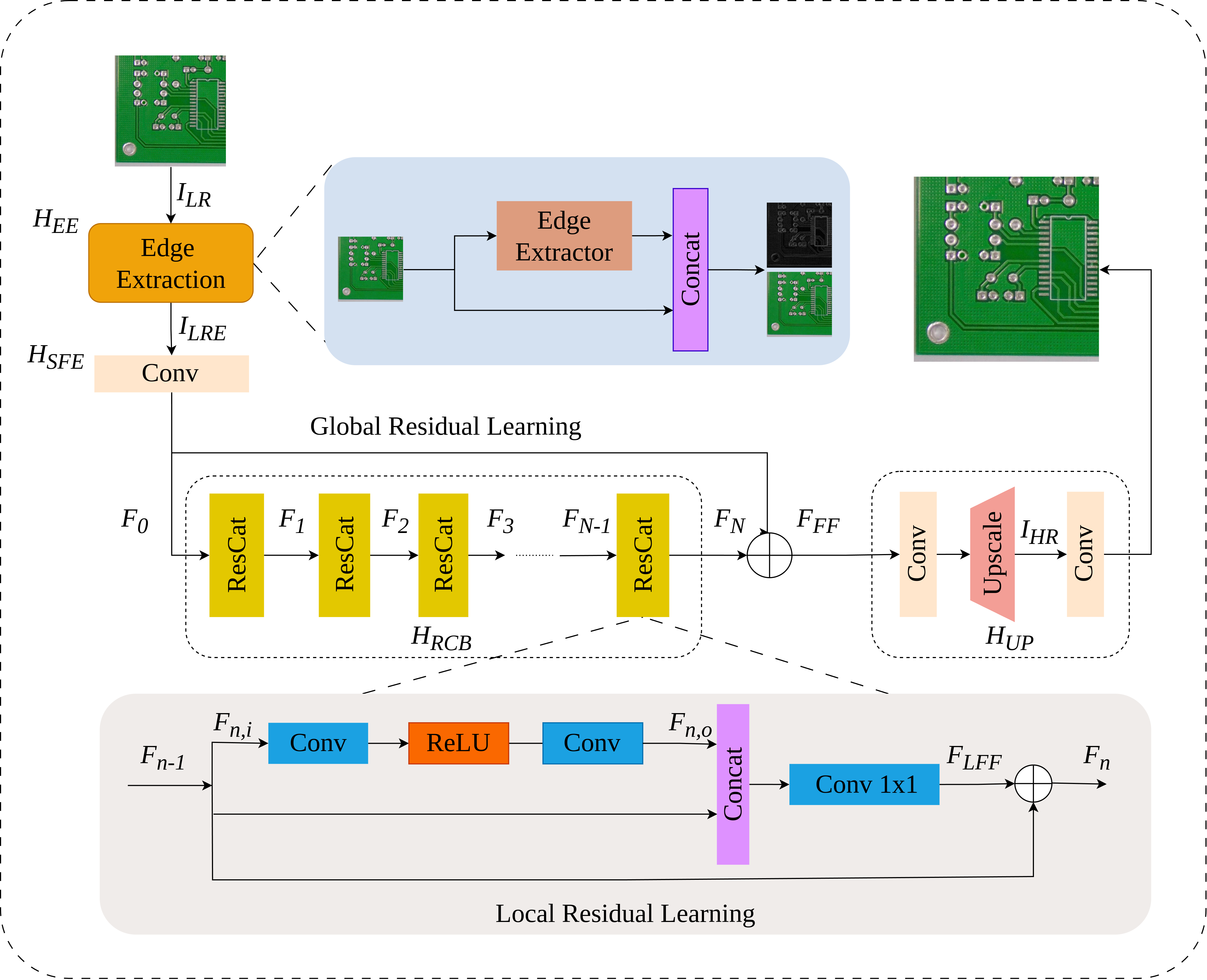}
    \caption{The architecture of the proposed ESRPCB Network.}
    \label{fig:esrpcbmodel}
\end{figure*}

The proposed ESRPCB network architecture illustrated in Fig.~\ref{fig:esrpcbmodel}, consists of primary components: edge extraction (EE), shallow feature extraction (SFE), residual concatenation blocks (RCB), and final up-sampling block (UP). The input and output of the ESRPCB model are denoted as $I_{LR}$ (low-resolution input) and $I_{HR}$ (high-resolution output), respectively.

Firstly, the input $I_{LR}$ undergoes an edge extraction process, yielding $I_{LRE}$:
\begin{equation}
    I_{LRE} = H_{EE} (I_{LR})
\end{equation}
where $H_{EE}$ represents the edge extraction process. Details of the edge extraction process, using Canny and Sobel algorithms, are elaborated in Section~\ref{3.2 edge extract}. The extracted edge features, $I_{LRE}$, enrich the subsequent feature extraction and global residual learning. 

The enhanced $I_{LRE}$ is processed through the shallow feature extraction module to generate initial features $F_0$:

\begin{equation}
    F_0 = H_{SFE}(I_{LRE})
\end{equation}
where $H_{SFE}$ denotes the convolution operation, which serves as the input to the ResCat. 

The network employs $N$ ResCat blocks for hierarchical feature extraction, with the output of the $N-th$ block denoted as $F_N$:

\begin{equation}
\begin{aligned}
    F_N &= H_{RCB,N}(F_{N-1}) \\
        &= H_{RCB,N}\left(H_{RCB,N-1}\left(...\left(H_{RCB,1}(F_0)\right)...\right)\right)
\end{aligned}
\end{equation}
where $H_{RCB, N}$ represents the operations of the $N-th$ ResCat block. These blocks preserve learned features across layers, leveraging their additive and concatenation skip connections to maintain feature integrity and avoid vanishing gradients. Detailed operations of the RCBs are explained in Section \ref{3.1.2 rescat}. 

The extracted hierarchical features undergo feature fusion through a global residual learning process:
\begin{equation}
    F_{FF}= F_N+F_0
\end{equation}
where $F_{FF}$ denotes the fusion operation, and "+" denotes the add skip connection. This ensures that both low-level and high-level features are effectively utilized.

The up-sampling (UP) module operates in the high-resolution space, reconstructing the output image $I_{HR}$:
\begin{equation}
    I_{HR} = H_{UP} (F_{FF})
\end{equation}
where $H_{UP}$ represents the up-sampling block. The up-sampling block includes a convolutional layer, followed by a Pixel Shuffle layer for resolution enhancement, and concludes with a final convolutional layer. 
Finally, the proposed ESRPCB model produces a comprehensive transformation of LR images into HR images by
\begin{equation}
    I_{HR} = H_{ESRPCB} (I_{LR})
\end{equation}
where $H_{ESRPCB}$ represents the complete ESRPCB network. 

\subsection{Edge extraction and ultilization} \label{3.2 edge extract}

Preserving edge properties is crucial for super-resolution algorithms, especially in applications such as PCB defect detection, where tiny defects are often defined by subtle structural details. Edge detection not only enhances the reconstruction quality but also provides essential context for distinguishing defects from noise. In the proposed ESRPCB framework, two edge detection algorithms, Sobel and Canny, are utilized to extract edge features that are then concatenated with the input image. The combined input, $\mathbf{X}_{\text{input}}$, is formulated as:

\begin{equation}
\mathbf{X}_{\text{input}} = \mathbf{E}_{\text{edge}} \oplus \mathbf{I}_{\text{image}}
\end{equation}
where $\mathbf{X}_{\text{input}}$ is input to the ESRPCB model, $\mathbf{E}_{\text{edge}}$ is extracted edge information, $\mathbf{I}_{\text{image}}$ is the original low-resolution image, and the symbol "$\oplus$" denotes the concatenation in the channel dimension.

Firstly, the Sobel algorithm is applied to our model, which is a gradient-based edge detection technique that calculates the gradient of image intensity at each pixel, emphasizing regions of rapid intensity change, which are typically edges. The Sobel filter uses two convolution kernels to compute gradients in horizontal ($G_x$) and vertical ($G_y$) directions:

    \begin{equation}
    G_x = \begin{bmatrix} 
      -1 & 0 & 1 \\
      -2 & 0 & 2 \\
      -1 & 0 & 1 
    \end{bmatrix}, 
    G_y = \begin{bmatrix} 
      -1 & -2 & -1 \\
      0 & 0 & 0 \\
      1 & 2 & 1 
    \end{bmatrix}
    \end{equation}
    
    The gradient magnitude is calculated as:
    \begin{equation}
        G = \sqrt{G_x^2 + G_y^2}
    \end{equation}

    Sobel emphasizes edge orientation and provides a clear structural representation, making it suitable for highlighting detailed edge information in PCB defect detection.
    
    Secondly, the Canny algorithm, a robust multi-stage edge detection method, is employed to preserve critical edge details while suppressing noise. The process involves the following steps:
    
    \begin{enumerate}
        \item \textit{Noise Reduction}: The Gaussian filter smoothens the image to reduce noise:
        \begin{equation}
        G(x, y) = \frac{1}{2\pi \sigma^2} \exp \left( -\frac{x^2 + y^2}{2\sigma^2} \right)
        \end{equation}
        where $\sigma$ is the Gaussian standard deviation. The smoothed image $I$ is:
        \begin{equation}
        I_{smooth} = I * G
        \end{equation}
        
        \item \textit{Gradient Calculation}: Gradients $G_x$ and $G_y$ are computed and the gradient magnitude and direction are derived as:
        \begin{equation}
        G_x = \frac{\partial I_{smooth}}{\partial x}, \quad G_y = \frac{\partial I_{smooth}}{\partial y}
        \end{equation}
        The magnitude and direction of the gradient are given by:
        \begin{equation}
        G = \sqrt{G_x^2 + G_y^2}, \quad \theta = \tan^{-1}\left(\frac{G_y}{G_x}\right)
        \end{equation}
        
        \item \textit{Non-maximum Suppression}: Gradients are thinned to preserve only the most significant edges along the direction $\theta$. This step refines the edges and thins them out.
        
        \item \textit{Double Thresholding}: Pixels are classified based on thresholds $T_{low}$ and $T_{high}$ with three categories: strong edges, weak edges, and non-edges. 
        \begin{itemize}
            \item Strong edges: $G > T_{high}$
            \item Weak edges: $T_{low} \leq G \leq T_{high}$
            \item Non-edges: $G < T_{low}$
        \end{itemize}
        
        \item \textit{Edge Tracking by Hysteresis}: Weak edges that are connected to strong edges are preserved, while isolated weak edges are discarded. This step finalizes the edge detection by linking edge segments together.
    \end{enumerate}
In this study, the optimization of Canny edge detector parameters, particularly the low and high thresholds, was performed empirically by conducting a series of experiments with different pairs of threshold values: $100 - 200$, $100 - 220$, $80 - 200$, and $80 - 220$. The qualitative results, presented in Table \ref{tab:qualitative_canny} indicate that the threshold pair 100-200 achieved the best performance in terms of Peak Signal-to-Noise Ratio (PSNR), Structural Similarity Index (SSIM), and detection accuracy, suggesting a balanced preservation of edge information and noise suppression. This pair was, therefore, adopted in the proposed $ESRPCB_C$ model.

\begin{table}[htbp]
\centering
\caption{Qualitative results of PSNR/SSIM for ESRPCB\_C model with various pairs of low and high threshold. Red highlights the best thresholds.}
\label{tab:qualitative_canny}

\begin{tabular}{|l|c|c|c|c|}
\hline
Type of defect   & 100 / 200      & 100 / 220      & 80 / 200       & 80 / 220       \\ \hline
mouse bite       & 30.47 / 0.8465 & 30.50 / 0.8444 & 30.48 / 0.8447 & 30.52 / 0.8453 \\
spur             & 30.75 / 0.8527 & 30.70 / 0.8496 & 30.64 / 0.8491 & 30.67 / 0.8497 \\
missing hole     & 30.47 / 0.8465 & 30.45 / 0.8441 & 30.41 / 0.8439 & 30.44 / 0.8448 \\
short            & 30.43 / 0.8424 & 30.37 / 0.8393 & 30.32 / 0.8388 & 30.39 / 0.8403 \\
open circuit     & 30.64 / 0.8505 & 30.61 / 0.8482 & 30.55 / 0.8477 & 30.60 / 0.8486 \\
spurious copper  & 30.36 / 0.8457 & 30.36 / 0.8429 & 30.30 / 0.8425 & 30.35 / 0.8433 \\ \hline
\textbf{Average} & \textcolor{red}{30.54} / \textcolor{red}{0.8476}& 30.50 / 0.8447 & 30.30 / 0.8425 & 30.50 / 0.8453 \\ \hline
\end{tabular}
\end{table}

The extracted edge features using Sobel and Canny algorithms are integrated into the ESRPCB framework as additional inputs, enriching the model's ability to retain structural details. This enhancement improves the super-resolution process, resulting in higher detection accuracy for tiny defects. 

\subsection{Residual Concatenation}\label{3.1.2 rescat}
\begin{figure}[htbp]
    \captionsetup[subfloat]{labelformat=empty}
    \begin{center}
        \includegraphics[width=0.8\textwidth]{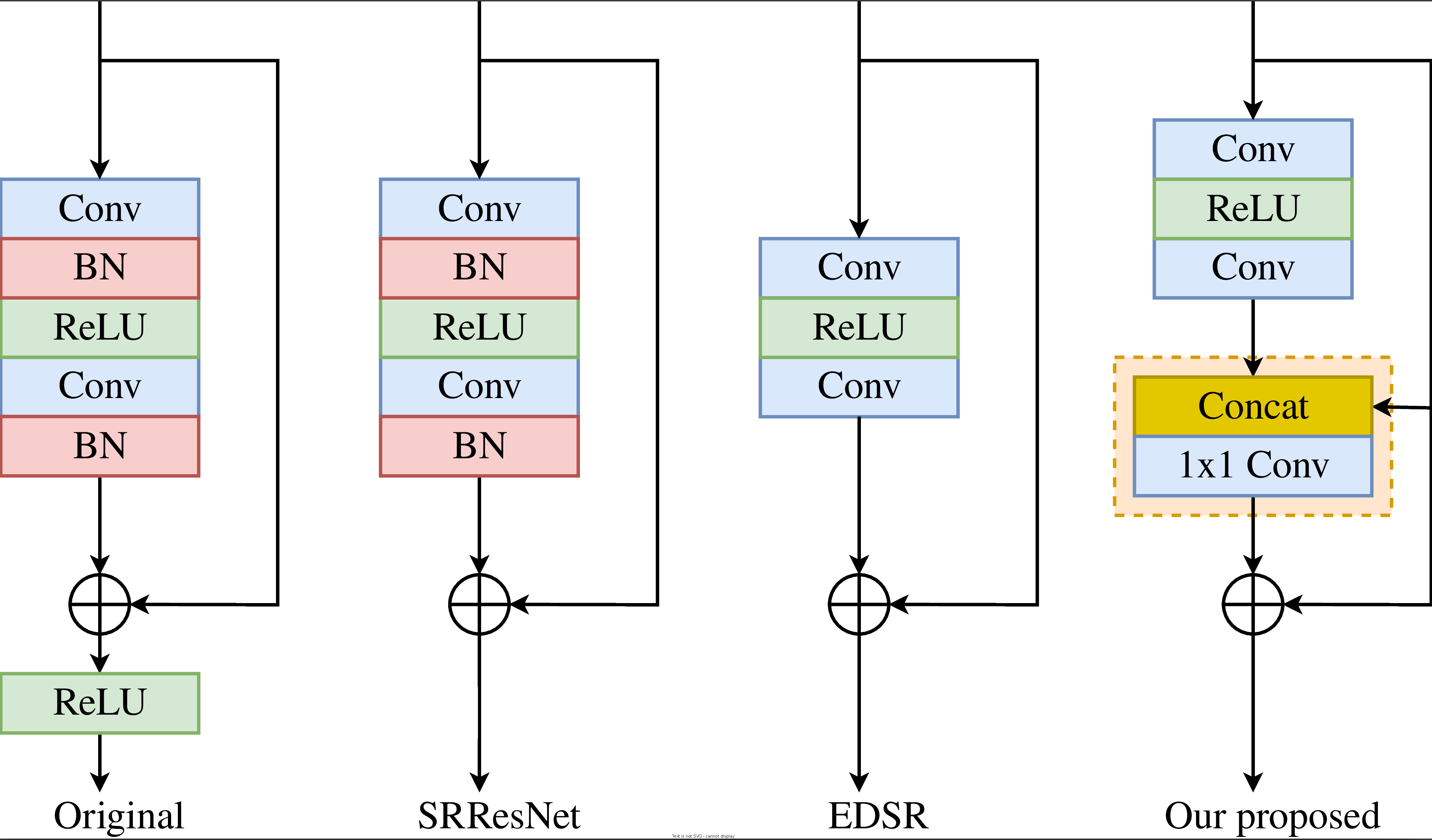}
    \end{center}
    \vspace*{-0.1cm}
    \captionsetup{justification=raggedright, singlelinecheck=false}
    \caption{The comparison of residual blocks in original ResNet~(\cite{he2016deep}), SRResNet~(\cite{ledig2017photo}), EDSR~(\cite{lim2017enhanced}) and our ResCat.}
    \label{fig:compare res block}
\end{figure}

Fig.~\ref{fig:compare res block} illustrates a comparison of the residual building blocks from the original ResNet~(\cite{he2016deep}), SRResNet~(\cite{ledig2017photo}), EDSR~(\cite{lim2017enhanced}), and the proposed ResCat block. Similar to EDSR, the ResCat block removes batch normalization layers to prevent the standardization of features, which can limit the model’s range flexibility. The ResCat block introduces a refined structure comprising Local Feature Extraction, Local Feature Fusion, and Local Residual Learning.

Local feature extraction (LFE) applies the input $F_{n,i}$ through a Convolutional (Conv) layer, a Rectified Linear Unit (ReLU) function, and a final Convolutional layer:
\begin{equation}
    F_{n,o}=H_{LFE}(F_{n,i})
\end{equation}
where $H_{LFE}$ represents LFE function. This step extracts hierarchical local features, enabling the network to capture intricate patterns from the input.

The extracted features $F_{n,o}$ and input $F_{n,i}$ are combined using a concatenation skip connection, followed by a $1\times1$ Convolutional layer to reduce dimensionality:

\begin{equation}
    F_{LFF} = H_{LFF}(F_{n,i},F_{n,o})
\end{equation}
where $H_{LFF}$ denotes the concatenation and dimensionality reduction operation. The LFF step ensures that all learned features from different paths are integrated effectively.

Local residual learning enhances the flow of information through the network by incorporating additive skip connections:

\begin{equation}
    F_{n}=F_{LFF}+F_{n,i}
\end{equation}
This step preserves the original input features, mitigates vanishing gradients, and ensures efficient gradient flow during backpropagation.

The ResCat block employs both additive and concatenation skip connections, leveraging the strengths of each. While additive skip connections maintain tensor sizes and ensure efficient gradient propagation, concatenation skip connections allow the model to retain features from different paths. This dual mechanism enables the network to capture diverse feature representations, improving reconstruction accuracy and robustness. By integrating these mechanisms, the ResCat block enhances feature extraction and fusion, ensuring that the model effectively learns intricate details while maintaining computational efficiency. This architecture plays a pivotal role in the superior performance of the ESRPCB framework for PCB defect detection.

\subsection{Ensemble-Based Defect Detection Model}
\begin{figure}[htbp]
    \centering
        \includegraphics[width=0.75\linewidth]{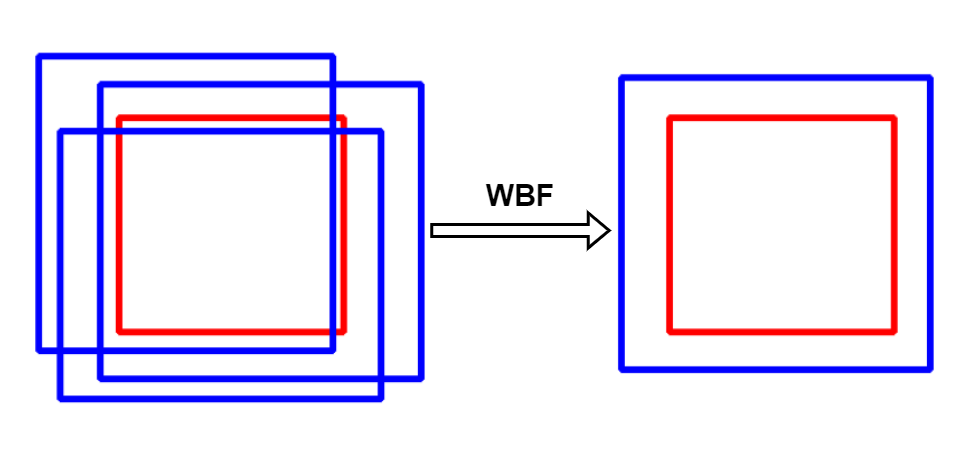}
    \caption{Schematic illustration for Weighted Boxes Fusion. Blue represents predictions from different models, while red represents the ground truth.}
    \label{fig:ensemble}
\end{figure}

To maximize the effectiveness of detecting defects in PCBs, an ensemble-based defect detection model is employed (Fig.~\ref{fig:ensemble}), combining predictions from the latest YOLO frameworks (YOLOv8 and YOLOv9) using Weighted Boxes Fusion (WBF) for output refinement. This ensemble approach leverages the complementary strengths of both YOLO models, improving detection accuracy and robustness.

YOLOv8 and YOLOv9 represent advanced iterations in the YOLO series, recognized for their real-time object detection capacities. YOLOv8~(\cite{yolov8}) introduces innovations such as optimized backbone and neck architecture, anchor-free detection heads, and an improved accuracy-speed trade-off. YOLOv9~(\cite{yolov9}), developed independently as an evolution of YOLOv5, integrated features like Programmable Gradient Information and the Generalized Efficient Layer Aggregation Network. These enhancements boost the model's learning capacity and preserve critical information throughout the detection process, achieving exceptional accuracy. 

To further refine the defect detection process, Weighted Boxes Fusion ~(\cite{weightboxfusion}) is integrated as a robust algorithm that combines predictions from multiple models. WBF improves detection reliability by merging bounding boxes from different models based on their confidence scores, mitigating inaccuracies from individual models. The steps of the WBF algorithm are as follows:

\begin{enumerate}
    \item \textit{Create list $B$:} Aggregate all predicted bounding boxes from multiple models into a single list, sorted by confidence scores $C$ in decreasing order.
    \item \textit{Initialize lists L and F:} List $L$ stores clusters of boxes, and list $F$ contains the fused boxes, one per cluster.
    \item \textit{Iterate through list $B$:} Match each box in $B$ to a cluster in $F$ using the  Intersection over Union (IoU) threshold. If no match is found, add the box to $L$ and $F$. Otherwise, add the box to the corresponding cluster.
    \item \textit{Box Fusion:} Recalculate the fused box's confidence score and coordinates:
    \begin{itemize}
        \item Confidence score:
            \begin{equation}
                C = \frac{\sum_{i=1}^{T} C_i}{T}
            \end{equation}
        \item Coordinates:
            \begin{align}
                X_{1,2} = \frac{\sum_{i=1}^{T} C_i \cdot X_{1,2_i}}{\sum_{i=1}^{T} C_i}\\
                Y_{1,2} = \frac{\sum_{i=1}^{T} C_i \cdot Y_{1,2_i}}{\sum_{i=1}^{T} C_i}
            \end{align}
                
        \end{itemize}
    \item \textit{Adjust Confidence Score:} Scale confidence scores based on the number of boxes in each cluster $T$ and the total number of models $N$:
    \begin{equation}
        C = C \times \frac{\min(T, N)}{N} \quad \text{or} \quad C = C \times \frac{T}{N}
    \end{equation}
\end{enumerate}

\section{Performance Evaluation} ~\label{sec:Eval}

\subsection{Dataset and Metrics} ~\label{sec:DatandMet}
\subsubsection{Datasets}
Extensive experiments were performed on the publicly available PCB defect dataset, which can be accessed at \url{http://robotics.pkusz.edu.cn/resources/dataset/} (\cite{pcbdataset}). This dataset contains 693 defective PCB images with corresponding annotation files. The average resolution of each image is $2777 \times 2138$ pixels, encompassing six defect types: missing hole, mouse bite, open circuit, short, spur, and spurious copper. Each image may include multiple defects. Table~\ref{tab:pcbdataset} provides detailed information on the dataset's distribution.
    
    \begin{table}[htbp]
        \centering
        \caption{The detail of PCB Defect Dataset.}
        \label{tab:pcbdataset}
        \begin{tabular*}{0.9\textwidth}{p{0.25\textwidth}>{\centering\arraybackslash}p{0.28\textwidth}
      >{\centering\arraybackslash}p{0.28\textwidth}}
        \toprule
        Type of defects & Number of images & Number of defects \\ \midrule
        missing hole    & 115              & 497               \\
        mouse bite      & 115              & 492               \\
        open circuit    & 116              & 482               \\
        short           & 116              & 491               \\
        spur            & 115              & 488               \\
        spurious copper & 116              & 503               \\ \midrule
        \textbf{Total}           & \textbf{693}              & \textbf{2953}              \\ \bottomrule
        \end{tabular*}
    \end{table}
    
For the experiments, \textcolor{blue} {an augmented version of the dataset presented in~(\cite{tddnet}) was used}. In this version, the original PCB images were systematically cropped into $600\times600$ pixels sub-images to expand the dataset to $10,668$ images as detailed in Table~\ref{tab:augpcbdataset}. The augmentation process was carefully designed to preserve the original defect distribution and prevent significant class imbalance. As reported in~(\cite{tddnet}), the defect distribution in the augmented dataset remains consistent with that of the original dataset, ensuring that the augmentation process did not introduce substantial bias.

\begin{table}[htbp]
\centering 
\caption{Overview of the Augmented PCB Defect Dataset.}
\label{tab:augpcbdataset}
\begin{tabular*}{0.9\textwidth}{p{0.25\textwidth}>{\centering\arraybackslash}p{0.28\textwidth}
>{\centering\arraybackslash}p{0.28\textwidth}}
\toprule
    Type of defects & Number of images & Number of defects \\ \midrule
    missing hole    & 1832             & 3612              \\
    mouse bite      & 1852             & 3684              \\
    open circuit    & 1740             & 3548              \\
    short           & 1732             & 3508              \\
    spur            & 1752             & 3636              \\
    spurious copper & 1760             & 3676              \\ \midrule
    \textbf{Total}           & \textbf{10668}           & \textbf{21664 }           \\ \hline
\end{tabular*}
\end{table}
    
\subsubsection{Metrics}
In this paper, the performance of the ESRPCB model is evaluated using PSNR and SSIM~\cite{wang2004image} for image quality measurement, while Mean Average Precision (mAP)~\cite{everingham2010pascal} at IoU = 0.50 is used as the metric for defect detection. PSNR quantifies the difference between the original and reconstructed images based on Mean Squared Error (MSE), whereas SSIM assesses structural similarity by considering three components: luminance, contrast, and structure. mAP evaluates defect detection accuracy by averaging precision across classes, with IoU $\geq$ 0.50 indicating correct detections. These metrics provide a comprehensive assessment of reconstruction and detection performance.

PSNR~\cite{wang2004image} is calculated using the Mean Squared Error (MSE)~\cite{wang2004image} between the original image \( I_1 \) and the reconstructed image \( I_2 \), as follows:

\begin{equation}
    \textit{MSE} = \frac{1}{mn} \sum_{i=1}^m \sum_{j=1}^n \left(I_1(i, j) - I_2(i, j) \right)^2
\end{equation}
where \(m\) and \(n\) represent the dimensions of the image, and \( I_1(i, j) \) and \( I_2(i, j) \) are the pixel values at position \( (i, j) \) in the original and reconstructed images, respectively. The PSNR is then computed as: 

\begin{equation}
    \textit{PSNR} = 10 \log_{10} \left(\frac{{\textit{MAX}^2}}{{\textit{MSE}}}\right)
\end{equation}
where \(\textit{MAX}\) is the maximum pixel value (typically 255 for 8-bit images).  

SSIM~\cite{wang2004image} is defined as:
\begin{equation}
\text{SSIM}(x, y) = \frac{(2\mu_x \mu_y + C_1)(2\sigma_{xy} + C_2)}{(\mu_x^2 + \mu_y^2 + C_1)(\sigma_x^2 + \sigma_y^2 + C_2)}
\end{equation}
where:
\begin{itemize}
    \item \(\mu_x, \mu_y\) are the mean pixel values of images \(x\) and \(y\),
    \item \(\sigma_x^2, \sigma_y^2\) are the variances of images \(x\) and \(y\),
    \item \(\sigma_{xy}\) is the covariance between images \(x\) and \(y\),
    \item \(C_1 = (K_1 \cdot L)^2\) and \(C_2 = (K_2 \cdot L)^2\) are constants to avoid division by zero, where \(L\) is the dynamic range of pixel values (e.g., 255 for 8-bit images), and \(K_1, K_2\) are small constants (e.g., \(K_1 = 0.01, K_2 = 0.03\)).
\end{itemize}

Defect detection performance was evaluated using the mAP50 metric, which measures the precision of bounding box predictions of bounding boxes. mAP50~\cite{everingham2010pascal} is calculated by averaging the Average Precision (AP) scores for all object classes at a fixed IoU threshold of 0.50. The process is represented below.

Precision:
\begin{equation}
P = \frac{TP}{TP + FP}
\end{equation}
Recall:
\begin{equation}
R = \frac{TP}{TP + FN}
\end{equation}
where:
\begin{itemize}
    \item TP: Number of correctly predicted objects.
    \item FP: Number of incorrectly predicted objects 
    \item FN: Number of actual objects that were missed.
\end{itemize}

Sort all predicted bounding boxes by confidence score. Compute Precision and Recall at different confidence thresholds to plot the Precision-Recall curve.

AP~\cite{everingham2010pascal} is calculated as the area under the Precision-Recall curve (AUC-PR):
\begin{equation}
AP = \int_{0}^{1} P(R) dR
\end{equation}
Finally, mAP50 is the mean of AP values across all object classes at IoU = 0.50:

\begin{equation}
    \textit{mAP}_{50} = \frac{1}{N} \sum_{i=1}^{N} \textit{AP}_{50}(i)
\end{equation}
where $N$ is the number of object classes.
\subsection{Super-Resolution Assessment} ~\label{sec:4.2 sr assessment}

For super-resolution assessment, the ESRPCB models were trained with RGB input patches of $(196 \times 196)$ pixels using the ADAM optimizer~(\cite{adam}) and MSE loss function to maximize PSNR. The models were trained for $300,000$ iterations with a learning rate initialized at $10^{-4}$, halved every 100,000 mini-batch updates, $\beta_1=0.9$, $\beta_2=0.999$, $\epsilon=10^{-8}$ and mini-batch size to 16. ESRPCB was compared with state-of-the-art (SOTA) models, including SRCNN (Super-Resolution Convolutional Neural Network)~(\cite{srcnn}), VDSR (Very Deep Super-Resolution Network)~(\cite{kim2016accurate}), SRResNet (Super-Resolution Residual Network)~(\cite{ledig2017photo}), and EDSR~(\cite{lim2017enhanced}), under identical training conditions on an NVIDIA RTX 2080 Ti GPU.

\begin{figure*}[htbp]
    \centering
    \includegraphics[width=\textwidth]{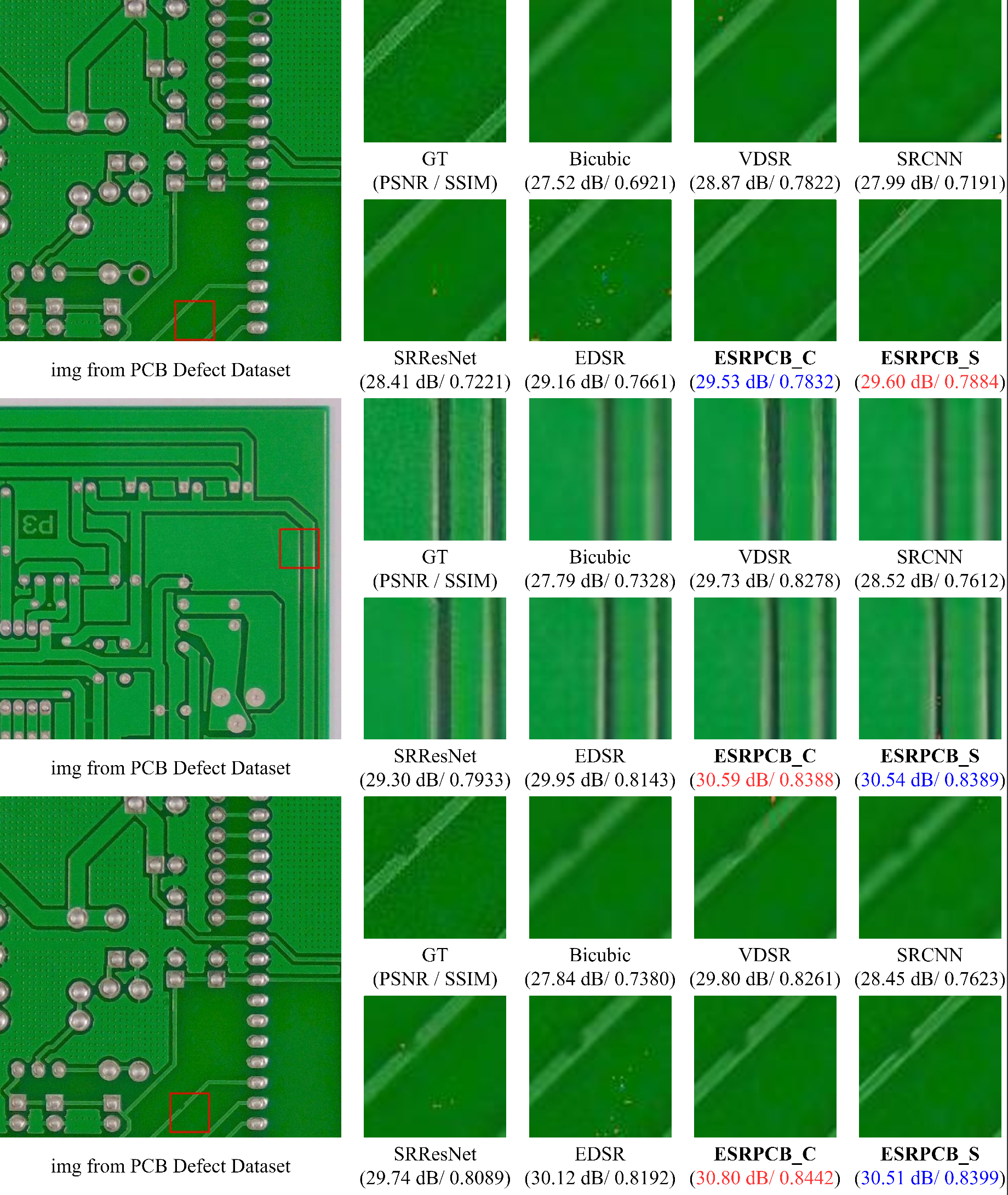}
    \caption{Visual comparison of super-resolution results for various models on $\times 4$ scale. The ground truth and reconstructed images are compared across metrics (PSNR/SSIM). Red highlights the best performance, and blue indicates the second-best.}
    \label{visual_result_1}
\end{figure*}

The results in Table~\ref{tab:sr results} and Figure~\ref{visual_result_1} clearly indicate that the ESRPCB models, particularly ESRPCB\_C and ESRPCB\_S, achieved superior performance compared to other state-of-the-art models in terms of both PSNR and SSIM. Specifically, ESRPCB\_C attained an average PSNR of $30.54$ dB, SSIM of $0.8476$, surpassing EDSR, the next-best model, by $0.42$ dB. Although the incremental improvement in PSNR might appear modest, it is significant for tasks requiring high precision, such as defect detection in tiny and complex regions of PCBs. The integration of edge-guided features effectively enhanced the reconstruction of fine details and minimized artifacts, thereby substantially improving the clarity and quality of the restored images. This enhancement is particularly critical for defect detection models, as higher-quality inputs directly contribute to more accurate and reliable defect identification.

\begin{table}[htbp]
    \centering
    \caption{Quantitative Comparison of Average PSNR and SSIM for Super-Resolution Models.}
    \label{tab:sr results}
    \begin{tabular*}{1.07\textwidth}{%
      p{0.13\textwidth}
      >{\centering\arraybackslash}p{0.1\textwidth}
      >{\centering\arraybackslash}p{0.1\textwidth}
      >{\centering\arraybackslash}p{0.1\textwidth}
      >{\centering\arraybackslash}p{0.1\textwidth}
      >{\centering\arraybackslash}p{0.1\textwidth}
      >{\centering\arraybackslash}p{0.1\textwidth}
      >{\centering\arraybackslash}p{0.1\textwidth}
    }
    \hline
    Methods  & mouse bite     & spur           & missing hole   & short          & open circuit   & spurious copper& average\\ \toprule
    Bicubic  & 26.20 / 0.7050 & 26.37 / 0.7124 & 26.11 / 0.7030 & 26.16 / 0.6996 & 26.24 / 0.7087 & 26.08 / 0.7030 & 26.20 / 0.7053 \\
    SRCNN    & 28.83 / 0.8044 & 29.00 / 0.8094 & 28.74 / 0.8044 & 28.72 / 0.8091 & 28.86 / 0.8091 & 28.65 / 0.8022 & 28.80 / 0.8050 \\
    VDSR     & 29.59 / 0.8371 & 29.79 / 0.8330 & 29.50 / 0.8364 & 29.48 / 0.8405 & 29.66 / 0.8405 & 29.41 / 0.8345 & 29.57 / 0.8371 \\
    SRResNet & 29.06 / 0.7910 & 29.17 / 0.7841 & 29.16 / 0.7955 & 28.85 / 0.7841 & 29.28 / 0.8000 & 29.09 / 0.7957 & 29.10 / 0.7934 \\
    EDSR     & 30.13 / 0.8332 & 30.31 / 0.8383 & 30.06 / 0.8324 & 30.02 / 0.9285 & 30.22 / 0.8372 & 29.96 / 0.8309 & 30.12 / 0.8334 \\
    ESRPCB\_S& 30.56 / 0.8470 & 30.72 / 0.8516 & 30.46 / 0.8458 & 30.39 / 0.8410 & 30.60 / 0.8494 & 30.33 / 0.8447 & \textcolor{blue}{30.51} / \textcolor{blue}{0.8466} \\
    ESRPCB\_C& 30.57 / 0.8479 & 30.75 / 0.8527 & 30.47 / 0.8465 & 30.43 / 0.8424 & 30.64 / 0.8505 & 30.36 / 0.8457 &\textcolor{red}{30.54} / \textcolor{red}{0.8476} \\
    \bottomrule
    \end{tabular*}
\end{table}

\subsection{PCB Defect Detection Assessment} ~\label{sec:4.3 pcb dd assessment}
\begin{figure*}[!htbp]
    \centering
        \includegraphics[width=\textwidth]{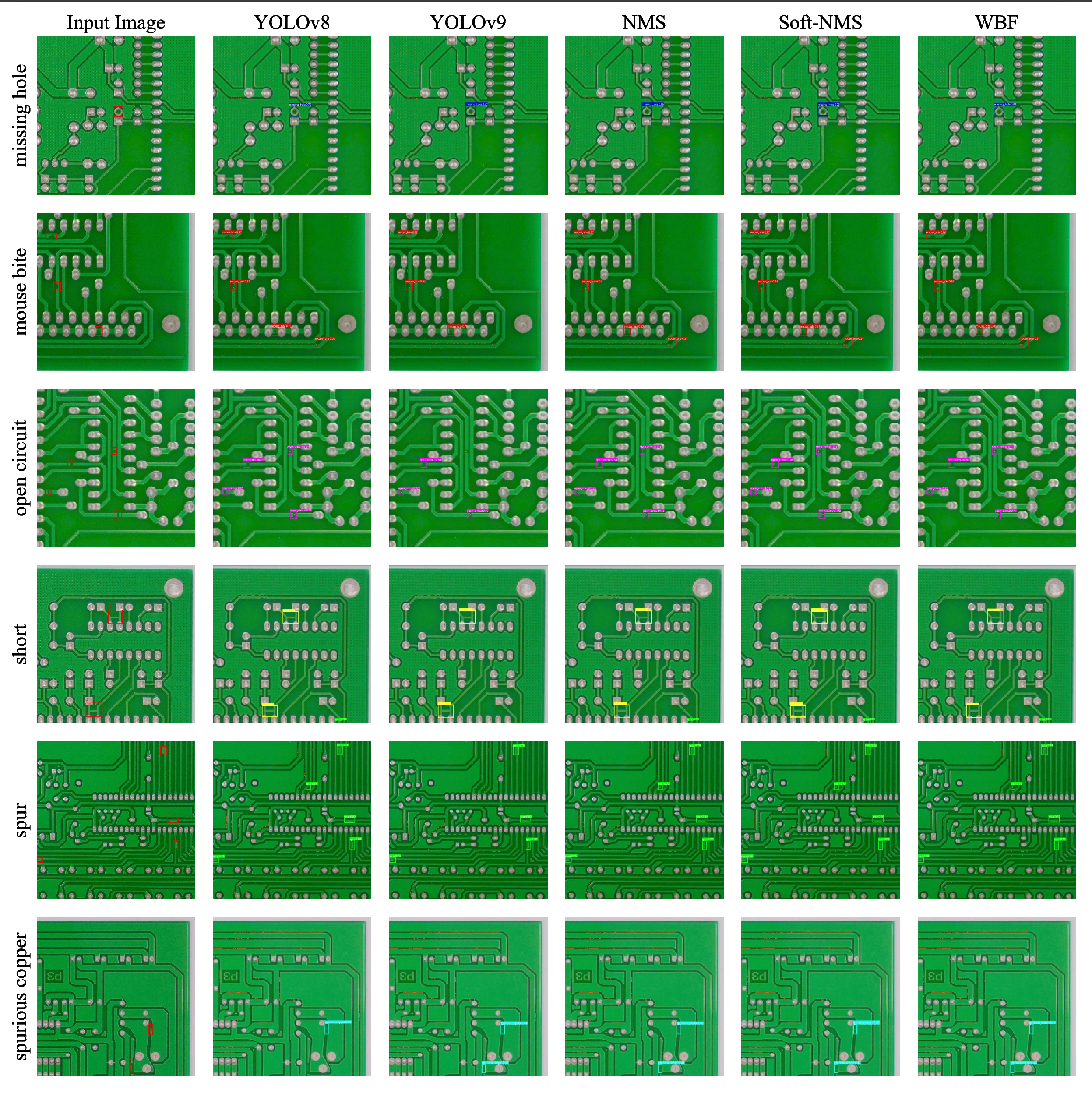}
    \caption{Defect detection results comparison.}
    \label{fig:defect_result}
\end{figure*}

In defect detection assessment, the outputs of the super-resolution models were evaluated using YOLOv8 and YOLOv9 object detection frameworks. The results in Table~\ref{tab:result detection} and Fig.~\ref{fig:defect_result} show that ESRPCB models significantly improved detection accuracy compared to other super-resolution methods. Using YOLOv8, ESRPCB\_C achieved an average mAP${50}$ of $0.965$, outperforming EDSR's $0.905$ and the LR baseline's $0.460$. To further enhance performance, Weighted Boxes Fusion was employed as an ensemble technique, which aggregated predictions from YOLOv8 and YOLOv9. WBF consistently outperformed other fusion methods, such as Non-Maximum Suppression (NMS) and Soft-NMS~(\cite{softnms}), achieving a mAP50 of $0.977$. This was only marginally lower than the ground truth performance (mAP50 of $0.987$), highlighting the high quality of reconstructed images and the robustness of the ensemble approach.

\begin{sidewaystable*}[!]
\centering
\begin{center}
\caption{Comparison of mAP$_{50}$ scores for Super-Resolution Models, red highlights the best performance, and bold values represent the ground truth (GT).}
\label{tab:result detection}
\begin{tabular}{clccccccccc}
\hline
Method & Type of defect  & LR    & Bicubic & SRCNN & VDSR  & SRResNet & EDSR  & ESRPCB\_S & ESRPCB\_C & GT    \\ \hline 
       & mouse bite      & 0.452 & 0.481   & 0.643 & 0.747 & 0.832    & 0.857 & 0.943   & 0.973   & \textbf{0.997} \\
       & spur            & 0.368 & 0.427   & 0.744 & 0.800 & 0.851    & 0.887 & 0.935   & 0.964   & \textbf{0.989} \\
       & missing hole    & 0.344 & 0.402   & 0.697 & 0.928 & 0.979    & 0.979 & 0.984   & 0.984   & \textbf{0.989} \\
YOLOv8 & short           & 0.608 & 0.646   & 0.834 & 0.788 & 0.888    & 0.958 & 0.944   & 0.963   & \textbf{0.978} \\
       & open circuit    & 0.619 & 0.678   & 0.866 & 0.813 & 0.889    & 0.915 & 0.965   & 0.984   & \textbf{0.990} \\
       & spurious copper & 0.371 & 0.409   & 0.686 & 0.733 & 0.831    & 0.837 & 0.912   & 0.921   & \textbf{0.976} \\
       & Average         & 0.460 & 0.507   & 0.745 & 0.801 & 0.878    & 0.905 & 0.947   & 0.965   & \textbf{0.987} \\ \hline 
       
       & mouse bite      & 0.533 & 0.559   & 0.703 & 0.753 & 0.846    & 0.905 & 0.952   & 0.975   & \textbf{0.989} \\ 
       & spur            & 0.475 & 0.513   & 0.776 & 0.815 & 0.860    & 0.915 & 0.950   & 0.970   & \textbf{0.989} \\ 
       & missing hole    & 0.344 & 0.420   & 0.767 & 0.947 & 0.979    & 0.989 & 0.993   & 0.993   & \textbf{0.989} \\ 
YOLOv9 & short           & 0.640 & 0.649   & 0.817 & 0.786 & 0.898    & 0.858 & 0.945   & 0.965   & \textbf{0.979} \\ 
       & open circuit    & 0.660 & 0.676   & 0.868 & 0.814 & 0.889    & 0.908 & 0.964   & 0.985   & \textbf{0.990} \\ 
       & spurious copper & 0.436 & 0.449   & 0.681 & 0.734 & 0.848    & 0.858 & 0.921   & 0.930   & \textbf{0.987} \\ 
       & Average         & 0.515 & 0.544   & 0.769 & 0.808 & 0.887    & 0.922 & 0.954   & 0.970   & \textbf{0.987} \\ \hline 
       
       & mouse bite      & 0.564 & 0.586   & 0.705 & 0.757 & 0.859    & 0.910 & 0.952   & 0.983   & -\\
       & spur            & 0.496 & 0.529   & 0.774 & 0.805 & 0.847    & 0.906 & 0.956   & 0.973   & -\\ 
       & missing hole    & 0.373 & 0.448   & 0.802 & 0.939 & 0.974    & 0.984 & 0.993   & 0.993   & -\\ 
NMS    & short           & 0.665 & 0.664   & 0.820 & 0.799 & 0.893    & 0.954 & 0.947   & 0.966   & -\\ 
       & open circuit    & 0.663 & 0.679   & 0.872 & 0.807 & 0.894    & 0.902 & 0.972   & 0.984   & -\\ 
       & spurious copper & 0.442 & 0.462   & 0.693 & 0.748 & 0.848    & 0.870 & 0.920   & 0.939   & -\\ 
       & Average         & 0.534 & 0.561   & 0.778 & 0.809 & 0.884    & 0.921 & 0.957   & 0.973   & -\\ \hline 
       
       & mouse bite      & 0.573 & 0.599   & 0.718 & 0.760 & 0.856    & 0.918 & 0.951   & 0.982   & -\\ 
       & spur            & 0.508 & 0.537   & 0.778 & 0.809 & 0.843    & 0.903 & 0.956   & 0.972   & -\\ 
       & missing hole    & 0.377 & 0.461   & 0.818 & 0.945 & 0.974    & 0.982 & 0.992   & 0.992   & -\\ 
Soft-NMS   & short           & 0.671 & 0.669   & 0.825 & 0.795 & 0.891    & 0.953 & 0.946   & 0.966   & -\\ 
       & open circuit    & 0.667 & 0.681   & 0.876 & 0.811 & 0.901    & 0.908 & 0.971   & 0.983   & -\\ 
       & spurious copper & 0.455 & 0.474   & 0.697 & 0.752 & 0.853    & 0.869 & 0.919   & 0.938   & -\\ 
       & Average         & 0.542 & 0.570   & 0.785 & 0.812 & 0.886    & 0.922 & 0.956   & 0.972   & -\\ \hline 
       
       & mouse bite      & 0.594 & 0.623   & 0.746 & 0.779 & 0.860    & 0.928 & 0.959   & 0.989   & -\\ 
       & spur            & 0.538 & 0.562   & 0.803 & 0.822 & 0.866    & 0.923 & 0.966   & 0.978   & -\\ 
       & missing hole    & 0.399 & 0.473   & 0.839 & 0.948 & 0.975    & 0.983 & 0.999   & 0.999   & -\\ 
WBF    & short           & 0.684 & 0.683   & 0.830 & 0.806 & 0.894    & 0.964 & 0.950   & 0.970   & -\\ 
       & open circuit    & 0.685 & 0.706   & 0.881 & 0.827 & 0.903    & 0.911 & 0.969   & 0.989   & -\\ 
       & spurious copper & 0.478 & 0.499   & 0.717 & 0.765 & 0.863    & 0.881 & 0.927   & 0.936   & -\\ 
       & Average         & \textcolor{red}{0.563} & \textcolor{red}{0.591} & \textcolor{red}{0.803} & \textcolor{red}{0.825} & \textcolor{red}{0.894} & \textcolor{red}{0.932} & \textcolor{red}{0.962} & \textcolor{red}{0.977} & -\\ \hline
\end{tabular}
\end{center}
\end{sidewaystable*}

\subsection{Complexity Assessment} \label{sec 4.4 complexity assess}
Computational complexity was analyzed to ensure the practicality of the ESRPCB models. The proposed ResCat blocks, as detailed in Table~\ref{tab:compare block}, including filters, parameters, and Batch Normalization (BN), further optimized the architecture by balancing computational efficiency and feature retention. Table~\ref{tab:complexity compare} compares processing time, network parameters, and MACs across super-resolution models. ESRPCB models demonstrated competitive GPU and CPU processing times while maintaining efficient parameter usage, comparable to EDSR but with enhanced performance.

\begin{table}[htbp]
\centering
\caption{Comparison of Residual Blocks Across Different Models.}
\label{tab:compare block}
\begin{tabular*}{0.9\textwidth}{%
      p{0.2\textwidth}
      >{\centering\arraybackslash}p{0.12\textwidth}
      >{\centering\arraybackslash}p{0.12\textwidth}
      >{\centering\arraybackslash}p{0.12\textwidth}
      >{\centering\arraybackslash}p{0.2\textwidth}}
\toprule
Options       & Original & SRResNet & EDSR    & Our Proposed \\ \midrule
Filters       & 64       & 64       & 64      &64           \\ 
Parameters    & 82368    & 82368    & 73728  &82112        \\ 
Use BN layers & Yes      & Yes      & No      &No           \\ \bottomrule
\end{tabular*}
\end{table}

\begin{table}[htbp]
\centering
\caption{Computational complexity comparison of Super-Resolution Models.}
\label{tab:complexity compare}
\begin{tabular*}{1\textwidth}{%
      p{0.1\textwidth}
      >{\centering\arraybackslash}p{0.28\textwidth}
      >{\centering\arraybackslash}p{0.28\textwidth}
      >{\centering\arraybackslash}p{0.1\textwidth}
      >{\centering\arraybackslash}p{0.1\textwidth}}
\toprule
 Models & Time Process (GPU)& Time Process (CPU) & Params & MACs \\ \midrule
SRCNN    & 27ms & 169ms  & 69K   & 1.56G  \\ 
VDSR     & 59ms & 1267ms & 667K  & 15.06G \\ 
SRResNet & 40ms & 322ms  & 1547K & 50.47G \\ 
EDSR     & 24ms & 221ms  & 1515K & 44.64G \\ 
ESRPCB\_S  & 28ms & 233ms  & 1614K & 46.93G \\ 
ESRPCB\_C  & 29ms & 229ms  & 1613K & 46.88G \\ \bottomrule
\end{tabular*}
\end{table}

\subsection{Comparison with other state-of-the-art methods} \label{sec: comparison sota}
To comprehensively demonstrate the advantages of the proposed method, its effectiveness is evaluated through two key perspectives: super-resolution analysis and defect detection analysis.

\subsubsection{Super-resolution analysis}
Extensive experiments were conducted to evaluate the image reconstruction performance of the ESRPCB models against several SOTA super-resolution architectures, including SRCNN~\cite{srcnn}, VDSR~\cite{kim2016accurate}, SRResNet~\cite{ledig2017photo}, and EDSR~\cite{lim2017enhanced}. These methods were evaluated using the PCB Defect dataset at a $\times4$ scale, with PSNR and SSIM as the key metrics for comparison (shown in Table~\ref{tab:sr results}).

The ESRPCB models, ESRPCB\_C and ESRPCB\_S, demonstrated exceptional reconstruction capabilities, achieving PSNR values of $30.54$ dB and $30.51$ dB, respectively. These results are $0.42$ dB and $0.39$ dB higher than the third-best model, EDSR, highlighting the superior reconstruction ability of the ESRPCB models. This performance underscores the effectiveness of incorporating edge-guided information, which enables better preservation and enhancement of image details. The evaluation metrics conclusively show that the proposed models outperform existing methods, setting a new benchmark in the field.

\subsubsection{Defect detection analysis}
The effectiveness of ESRPCB in defect detection was evaluated by feeding reconstructed images into YOLOv8 and YOLOv9 object detection models. As shown in Table~\ref{tab:result detection}, ESRPCB methods outperformed all other SOTA super-resolution models in enhancing defect detection accuracy. For instance, using YOLOv8, ESRPCB\_C and ESRPCB\_S achieved mAP50 scores of $0.965$ and $0.947$, respectively, surpassing EDSR, which scored $0.905$.
\begin{figure}[htbp] 
    \captionsetup[subfloat]{labelformat=empty}
    \begin{center}
        \newcommand{\rowArg}{2.65cm}
        \newcommand{\patchsize}{6.7cm}
        \setlength\tabcolsep{6pt}
        \begin{tabular}{c c c}
            \subfloat[YOLOv8]{\includegraphics[width=\patchsize]{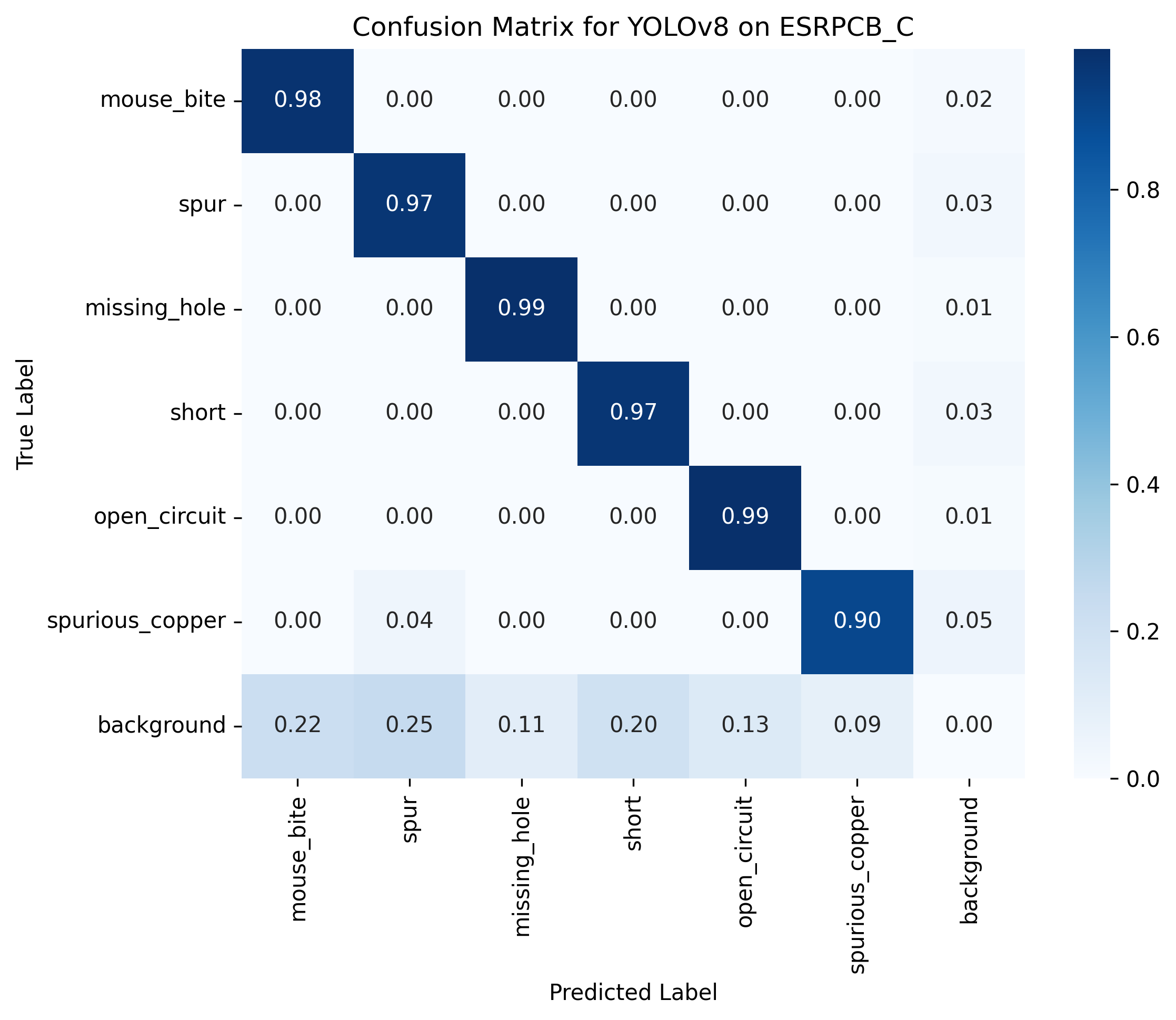}} &
            \subfloat[YOLOv9]{\includegraphics[width=\patchsize]{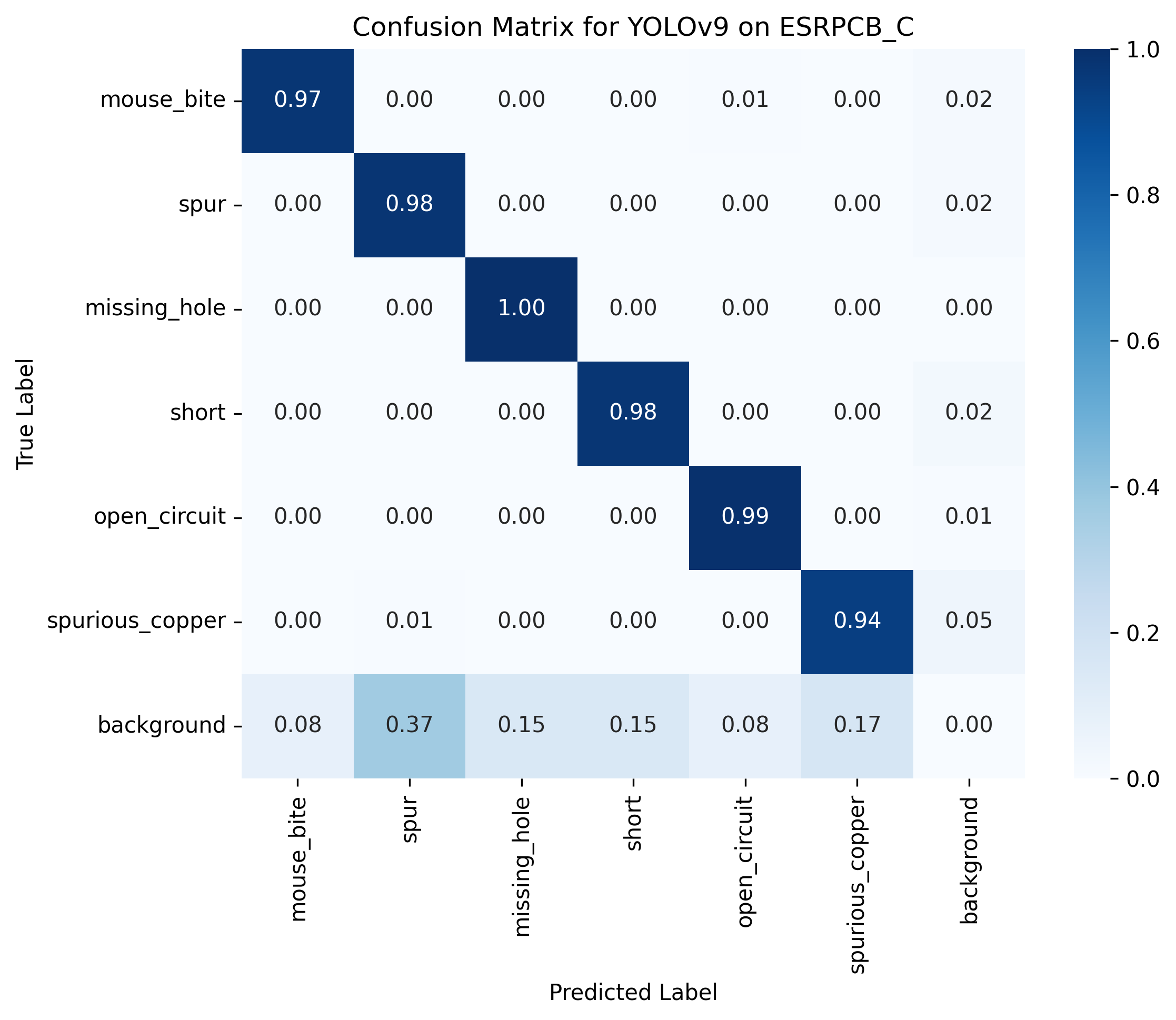}} \\ [0cm]
            \subfloat[WBF]{\includegraphics[width=\patchsize]{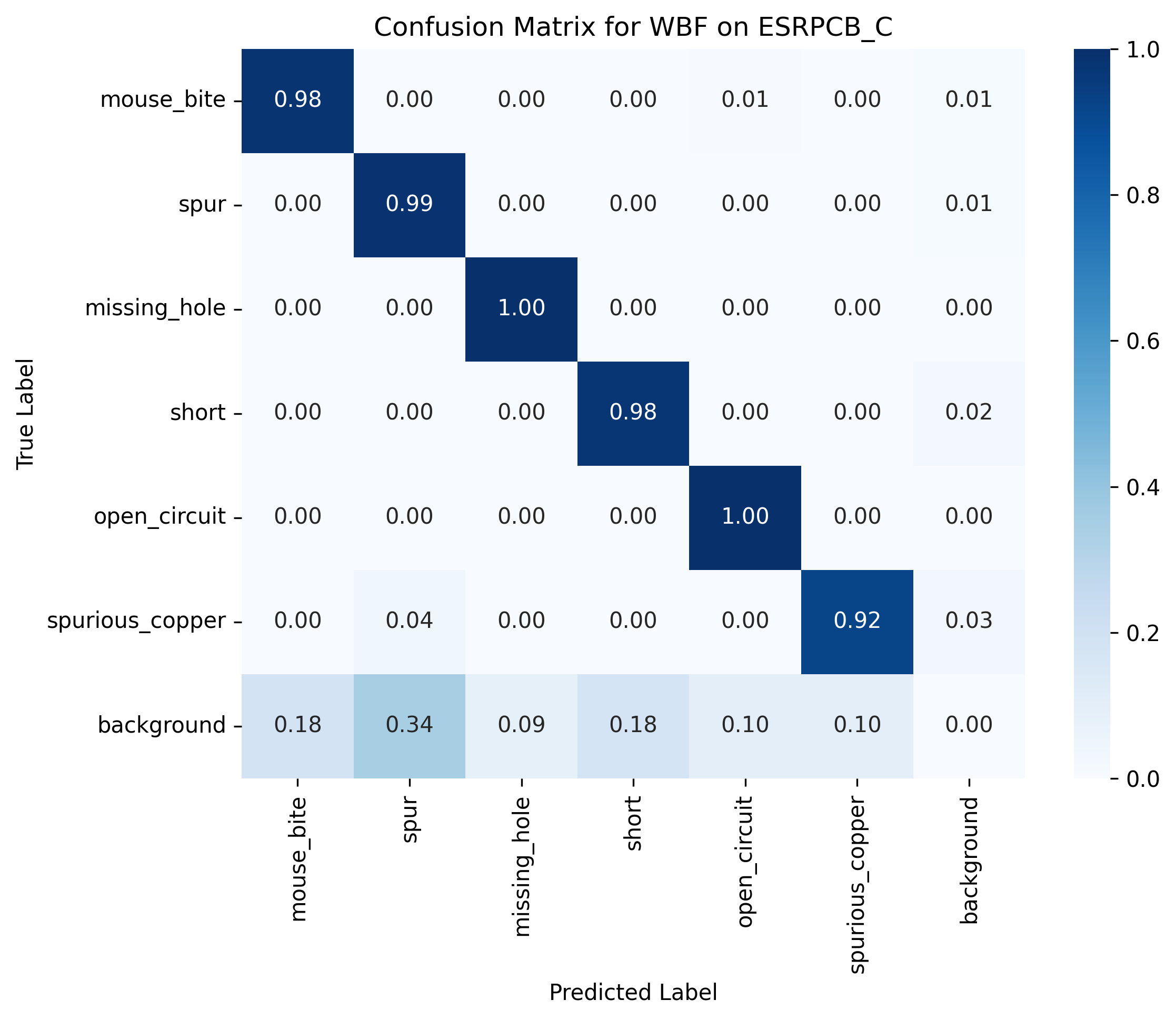}} &
            \subfloat[Soft-NMS]{\includegraphics[width=\patchsize]{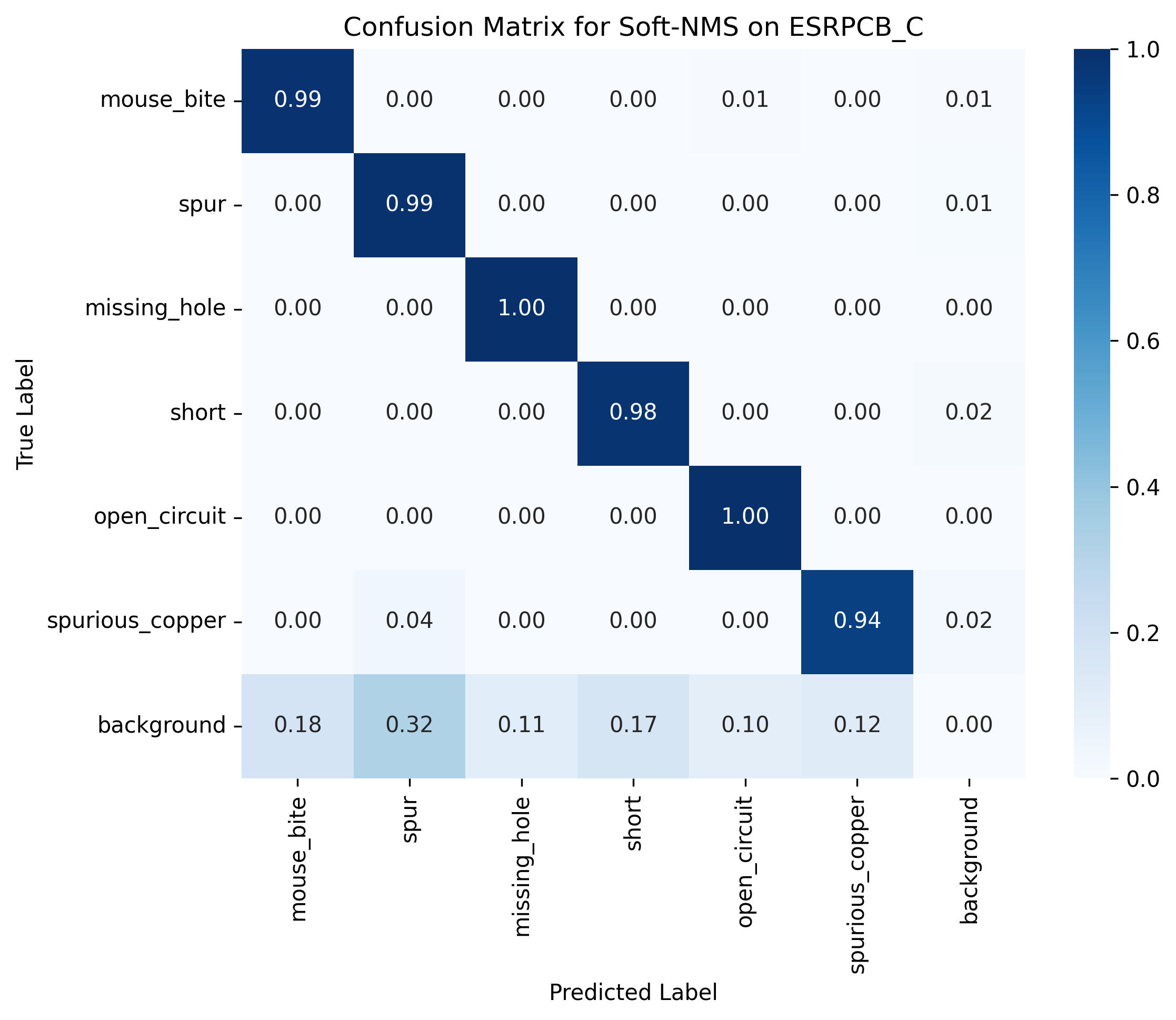}} 
        \end{tabular}
    \end{center}
    \vspace*{-0.3cm}
    \caption{Confusion Matrix for Defect detection models.}
    \label{fig:confusionmatrix}
\end{figure}

To provide a more in-depth analysis of the model's classification performance, confusion matrices are presented in Figure~\ref{fig:confusionmatrix}, which illustrate the distribution of true positives, false positives, and false negatives across defect categories. The results indicate that ESRPCB significantly reduces false negatives, particularly for missing hole and spurious copper defects, where fine-grained details are essential for accurate classification. The improved reconstruction quality enhances feature extraction, leading to better recognition accuracy in these challenging defect types.

Moreover, the integration of Weighted Boxes Fusion as an ensemble strategy further improved detection accuracy. By aggregating predictions from YOLOv8 and YOLOv9, WBF achieved an mAP$_{50}$ of $0.977$, outperforming individual models and alternative ensemble techniques such as Non-Maximum Suppression (NMS) and Soft-NMS. The confusion matrix for WBF demonstrates superior consistency in predictions, reducing both false positives and false negatives across all defect categories. These results highlight the robustness and effectiveness of ESRPCB in improving defect detection for tiny PCB defects, particularly in complex low-resolution scenarios.

\begin{figure}[htbp] 
    \captionsetup[subfloat]{labelformat=empty}
    \begin{center}
        \newcommand{\rowArg}{2.65cm}
        \newcommand{\patchsize}{6.7cm}
        \setlength\tabcolsep{6pt}
        \begin{tabular}{c c c}
            \subfloat[YOLOv8]{\includegraphics[width=\patchsize]{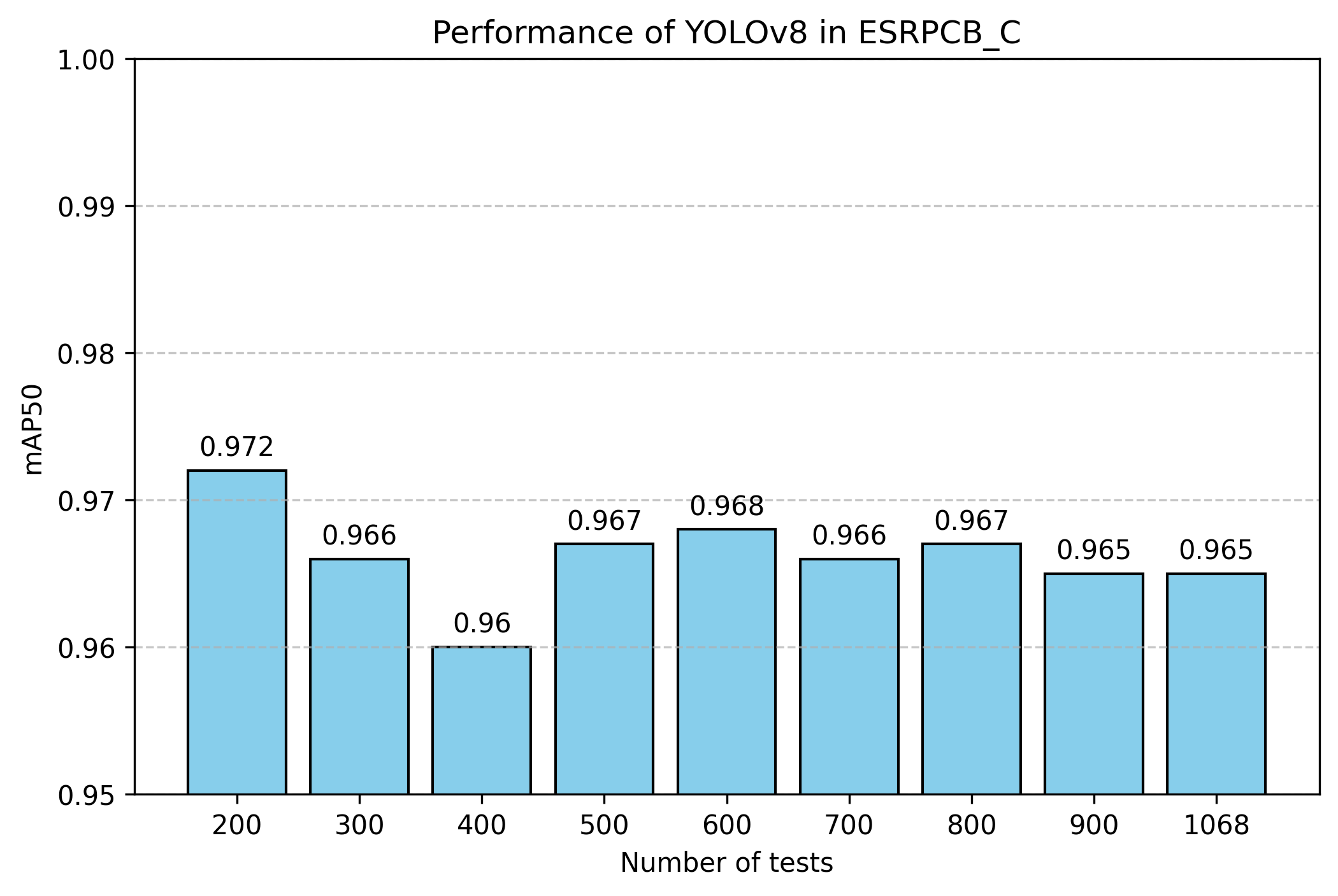}} &
            \subfloat[YOLOv9]{\includegraphics[width=\patchsize]{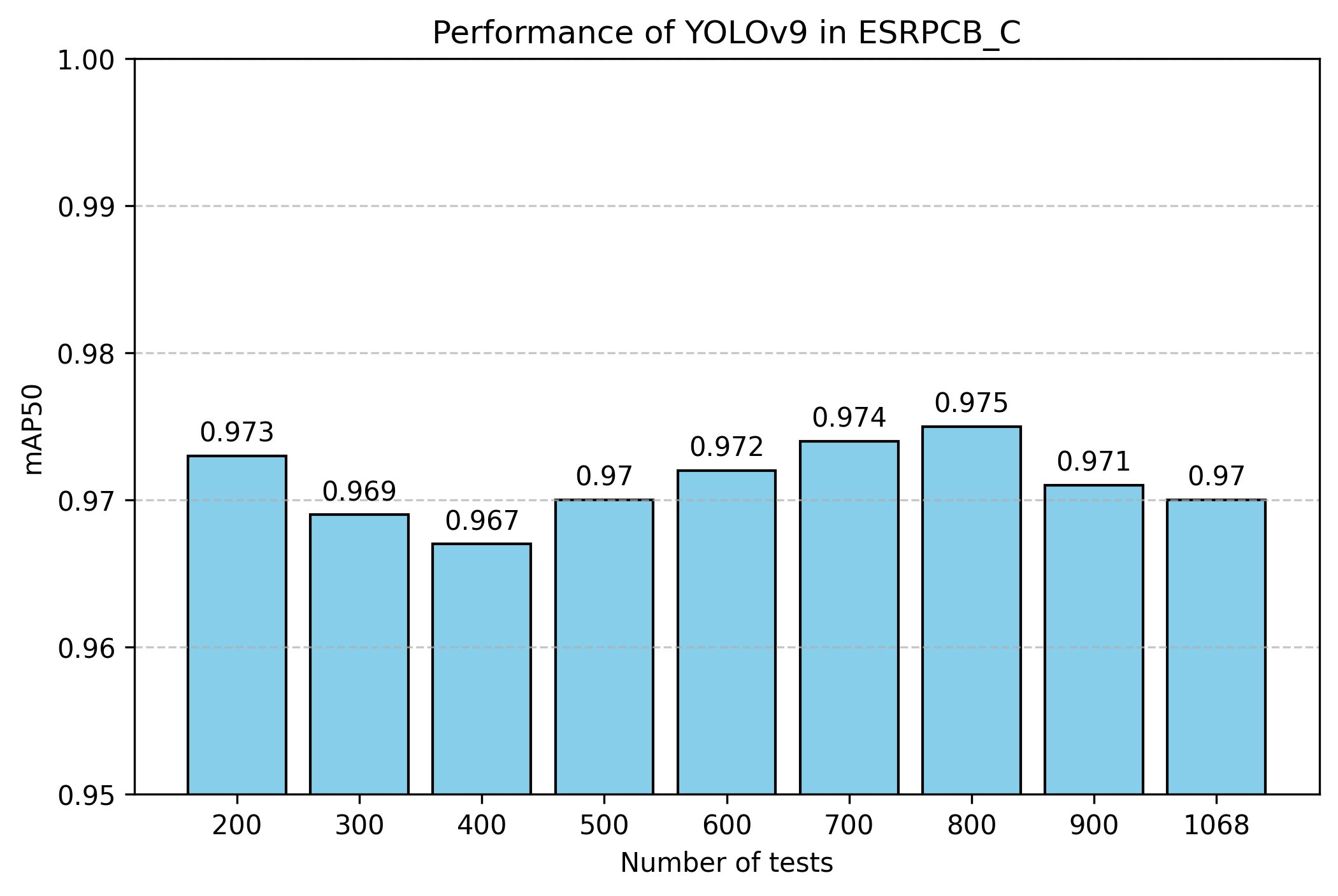}} \\ [0cm]
            \subfloat[Soft-NMS]{\includegraphics[width=\patchsize]{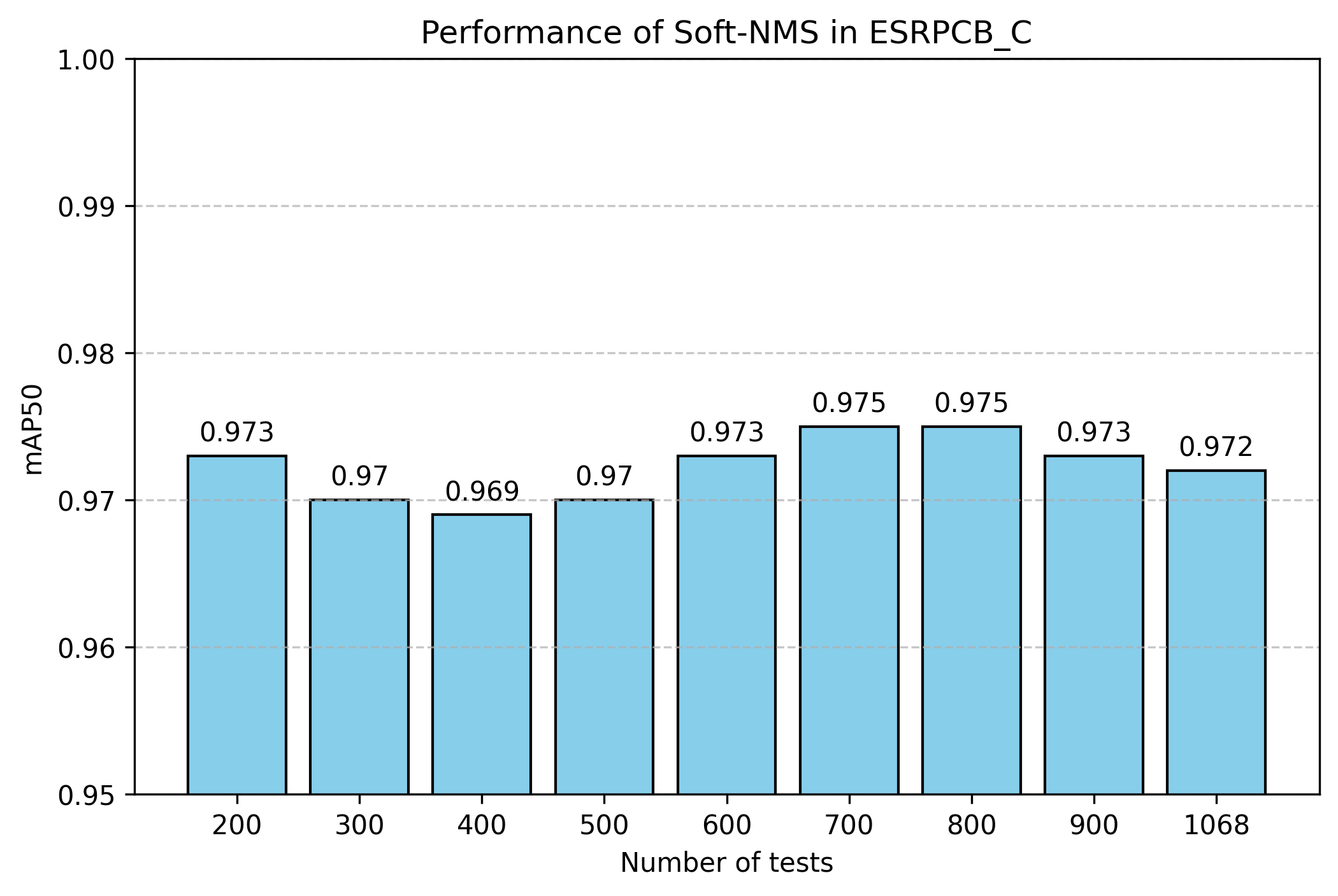}} &
            \subfloat[WBF]{\includegraphics[width=\patchsize]{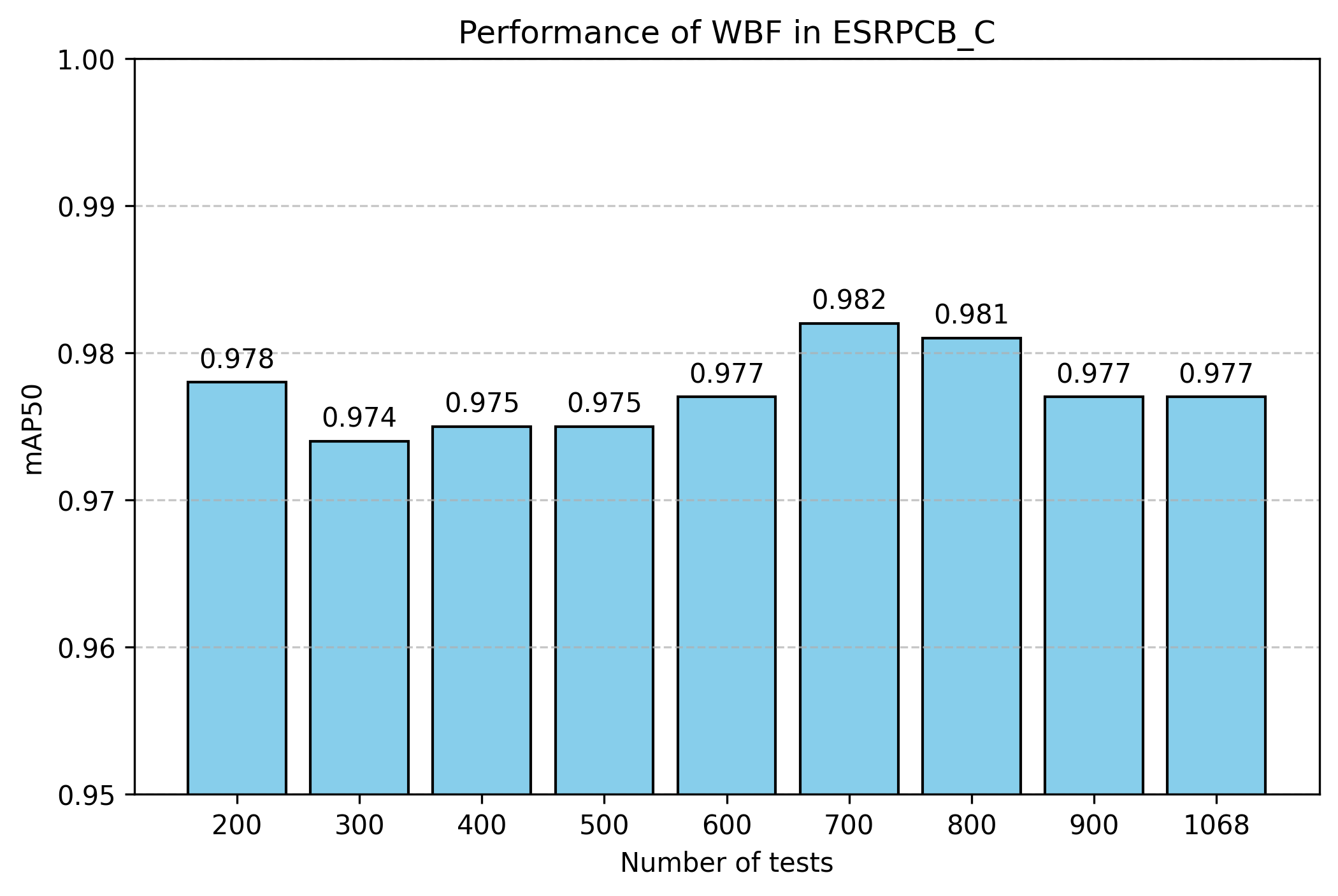}} 
        \end{tabular}
    \end{center}
    \vspace*{-0.3cm}
    \caption{Results of the Defect Detection model using varying dataset sizes.}
    \label{fig:defect_detection}
\end{figure}
To ensure a rigorous evaluation, the ESRPCB framework was tested across varying dataset sizes, ranging from 200 to 1068 images, in order to assess the model’s robustness under different data distributions. Rather than relying solely on a single performance metric, 95

To ensure a rigorous evaluation, the ESRPCB framework was tested across varying dataset sizes, ranging from 200 to 1068 images, to assess the robustness of the model across different data distributions.  Instead of relying on a single performance metric, $95\%$ confidence intervals (CI) were computed to determine the reliability of our reported mAP50 values. The results indicate that:
\begin{itemize}
    \item YOLOv8 on ESRPCB\_C achieved an average mAP50 of 0.9662 with a 95\% CI: [0.9638, 0.9686].
    \item YOLOv9 on ESRPCB\_C resulted in mAP50 of 0.9712 with a 95\% CI: [0.9693, 0.9732].
    \item Soft-NMS on ESRPCB\_C attained mAP50 of 0.9722 with a 95\% CI: [0.9706, 0.9739].
    \item WBF on ESRPCB\_C achieved the highest performance, with mAP50 of 0.9773 and a 95\% CI: [0.9753, 0.9794].
\end{itemize}

\section{Conclusion} ~\label{sec:conclusion}
In this paper, ESRPCB - a novel edge-guided super-resolution model with ResCat and ensemble learning for tiny PCB defect detection is presented. By integrating edge information and a ResCat structure into the Enhanced Deep Super-Resolution (EDSR) architecture, ESRPCB models demonstrated state-of-the-art performance in reconstructing high-fidelity images. Experimental results show that ESRPCB\_C and ESRPCB\_S achieved an average PSNR of $30.54$ dB and $30.51$ dB, respectively, surpassing the next-best model (EDSR) by $0.42$ dB. Similarly, the models improved SSIM from $0.8334$ (EDSR) to $0.8476$, demonstrating better structural preservation. In defect detection, ESRPCB models increased mAP50 from $0.905$ (EDSR) to $0.965$ (ESRPCB\_C) and $0.947$ (ESRPCB\_S), proving that the edge-guided SR framework significantly enhances defect recognition. Furthermore, the application of Weighted Boxes Fusion as an ensemble method in defect detection yielded significant accuracy improvements compared to individual models and alternative ensemble techniques. These improvements demonstrate that incorporating edge guidance and ensemble-based defect detection enhances PCB quality assurance, particularly in challenging imaging conditions where defects are minute and hard to discern.

Despite these promising results, several challenges and future directions remain in real-world deployment of super-resolution and defect detection models. One major challenge is balancing computational efficiency and model complexity. Future research could explore knowledge distillation and model compression techniques to reduce computational overhead while maintaining detection accuracy. Additionally, real-world industrial PCB inspection systems often deal with ultra-high-resolution images. Developing scalable models that can efficiently process such large-scale data while preserving detection accuracy remains an open research problem. Looking ahead, the developmental tendency in PCB defect detection is moving toward real-time, highly robust, and adaptive AI-driven solutions. Future research should focus on:
    \begin{itemize}
        \item Lightweight and efficient super-resolution models that balance high reconstruction quality with low computational cost.
        \item Few-shot and self-supervised learning approaches to enable robust defect detection with limited annotated data.
        \item Integration of Transformer-based models to capture global dependencies for improved defect localization.
        \item Multi-modal fusion techniques, such as combining optical imaging with X-ray or hyperspectral imaging, to detect defects beyond visual spectrum limitations.
        \item Edge AI deployment, where super-resolution and detection models are optimized for real-time inference on embedded and IoT devices for on-site quality inspection.
    \end{itemize}

\section*{Acknowledgement}
This research is funded by Vietnam National Foundation for Science and Technology Development (NAFOSTED) under grant number NCUD.02-2024.09.

\bibliographystyle{elsarticle-harv} 
\bibliography{ref}

\end{document}